\documentclass{article}

\usepackage{enumitem}
\usepackage{amsmath}

\usepackage[final]{neurips_2025}

\usepackage[utf8]{inputenc} 
\usepackage[T1]{fontenc}    
\usepackage{hyperref}       
\usepackage{url}            
\usepackage{booktabs}       
\usepackage{amsfonts}       
\usepackage{nicefrac}       
\usepackage{microtype}      
\usepackage{xcolor}         
\usepackage{wrapfig}
\usepackage{graphicx}
\usepackage[table]{xcolor}
\usepackage{multirow}

\usepackage{authblk}

\title{Multimodal 3D Genome Pre-training}

\author[1]{Minghao Yang}
\author[1]{Pengteng Li}
\author[2]{Yan Liang}
\author[1]{Qianyi Cai}
\author[3]{Zhihang Zheng}
\author[3]{Shichen Zhang}
\author[1]{Pengfei Zhang}
\author[4,*]{Zhi-An Huang}
\author[1,5,*]{Hui Xiong}

\affil[1]{Thrust of Artificial Intelligence, The Hong Kong University of Science and Technology (Guangzhou), China}
\affil[2]{School of Artificial Intelligence, South China Normal University, China}
\affil[3]{Thrust of Bioscience and Biomedical Engineering, The Hong Kong University of Science and Technology (Guangzhou), China}
\affil[4]{Department of Computer Science, City University of Hong Kong (Dongguan), China}
\affil[5]{Department of Computer Science and Engineering, The Hong Kong University of Science and Technology Hong Kong SAR, China}

\begin{document}

\maketitle

\begingroup
\def\thefootnote{*}\footnotetext{Correspondence to Zhi-An Huang <huang.za@cityu-dg.edu.cn>, Hui Xiong <xionghui@ust.hk>.}
\def\thefootnote{}\footnotetext{The source code of MIX-HIC is available at \url{https://github.com/myang998/MIX-HIC}.}
\endgroup

\vspace{-8mm}
\begin{abstract}
\label{sec:abstract}
Deep learning techniques have driven significant progress in various analytical tasks within 3D genomics in computational biology. However, a holistic understanding of 3D genomics knowledge remains underexplored. Here, we propose \textit{\textbf{MIX-HIC}}, the first multimodal foundation model of 3D genome that integrates both Hi-C contact maps and epigenomic tracks, which obtains unified and comprehensive semantics. For accurate heterogeneous semantic fusion, we design the cross-modal interaction and mapping blocks for robust unified representation, yielding the accurate aggregation of 3D genome knowledge. Besides, we introduce the first large-scale dataset comprising over \textit{\textbf{1 million}} pairwise samples of Hi-C contact maps and epigenomic tracks for high-quality pre-training, enabling the exploration of functional implications in 3D genomics. Extensive experiments show that MIX-HIC significantly surpasses existing state-of-the-art methods in diverse downstream tasks. This work provides a valuable resource for advancing 3D genomics research.
\end{abstract}

\vspace{-5mm}
\section{Introduction}
\vspace{-2mm}
\label{sec:intro}
The three-dimensional (3D) organization of chromosomes within the nucleus plays a pivotal role in gene regulation and cellular function \cite{zhang2024computational, monteagudo2024impact}. Key topological features of the 3D genome, such as chromatin loops that bring distant regulatory elements into close physical proximity with their target genes, are essential for cell-type-specific transcriptional regulation. High-resolution 3D chromatin interactions can be quantified through high-throughput chromosome conformation capture (Hi-C) technique \cite{lieberman2009comprehensive}. Understanding the mechanisms of how the 3D genome influences gene expression can unveil pivotal insights into cellular functionality, developmental biology, and disease mechanisms \cite{ferrer2024transcription}.


Recently, computational models have emerged as a powerful tool to unravel the intricate associations between the 3D chromatin structure, epigenome, and transcriptome. 
Existing approaches predict various genomic features, including 3D chromatin contact maps \cite{yang2023epiphany, tan2023cell, fudenberg2020predicting}, chromatin loops \cite{salameh2020supervised, matthey2020computer, wang2022dloopcaller}, and gene expression \cite{zhang2023generalizable, karbalayghareh2022chromatin}, often leveraging DNA sequences, Hi-C contact maps and epigenomic tracks.
Though successful, most of these methods are limited to a single specific task and struggle to integrate the diverse and heterogeneous information of the 3D genome, hindering a comprehensive understanding of its complex organization.



Recent progress in large-scale foundation models has demonstrated remarkable success in various fields of computational biology, such as molecular representations \cite{wang2024exploring, chang2024bidirectional}, medical imaging \cite{li2023llava, xia2024cares}, proteomics \cite{cheng2024training, li2024prosst}, and genomics \cite{icml2024vqdna, schiff2024caduceus}. 
Inspired by these advancements, we aim to develop a multimodal foundation model to address the above-mentioned limitations of analyzing 3D genomic downstream tasks in isolation.

Modeling the multimodal foundation model of 3D genome introduces three key challenges. 
First, Hi-C contact maps and epigenomic tracks have inherently distinct characteristics, making integration difficult. 
Simply aligning features from the two modalities and projecting them into a unified latent space would primarily capture modal-invariant knowledge like gene regulatory mechanisms, which are governed by both chromatin spatial organization and epigenomic tracks. 
However, this approach tends to overlook modal-specific characteristics, such as precise chemical modifications and chromatin states revealed by epigenomic tracks, which are essential factors for fine-grained 3D genome analysis.
This can lead to information loss and degrade downstream task performance (\textit{Refer to Appendix \ref{sec:Analysis} for theoretical analysis}).
Second, the unified representation from heterogeneous 3D genomic and epigenomic data must exhibit robust generalization capabilities; otherwise, the model may struggle to adapt effectively to diverse downstream tasks \textit{e.g.}, generation and regression.
Third, uncovering implicit semantic relationships between 3D genomic and epigenomic data is crucial for addressing the data scarcity problem in 3D genomics. Especially, the high experimental costs of Hi-C sequencing in real-world applications limit data accessibility, leading to incomplete representations and degraded model performance. Pre-training a model to learn implicit multimodal structural connections offers significant benefits for downstream tasks, particularly when only single-modality data is available. This enables the model to leverage structural knowledge to compensate for missing modality semantics, thereby enriching the overall data representation.

Hence, we introduce MIX-HIC, the first multimodal foundation model for 3D genomics to extract fine-grained knowledge from 3D genome and epigenomic tracks, enabling efficient adaptation to diverse downstream tasks with superior performance. 
Specifically, MIX-HIC first incorporates two distinct encoders to capture the refined features from 3D genome contact maps and epigenomic tracks, leveraging cell type-specific information for accurate predictions in novel cell types. 
To address the challenge of integrating heterogeneous data, we propose a cross-modal interaction block to capture both modal-invariant and modal-specific representations, regularized by contrastive learning and orthogonal constraints, preserving both shared and distinctive information across modalities. 
Additionally, a cross-modal mapping block facilitates information exchange between modalities, ensuring robust representation even with single-modality input. 
To comprehensively capture 3D genome knowledge, we have curated a large-scale dataset that consists of 1,275,948 pair samples of Hi-C contact maps and epigenomic tracks for rapid adaptation to downstream tasks through task-specific decoders. Notably, this is the largest paired dataset for the 3D genome analysis to date.
MIX-HIC is evaluated across diverse downstream tasks, demonstrating its effectiveness and robustness in comparison to other state-of-the-art methods. In summary, the main contributions are:

\begin{itemize}[leftmargin=10pt]
\item We propose \textbf{\textit{the first 3D genomic multimodal foundation model}}, integrating Hi-C contact maps and epigenomic tracks to establish a new paradigm for 3D genome analysis.
\item MIX-HIC features a novel architecture with two key components: (1) a cross-modal interaction block to capture both shared and unique biological patterns across modalities; and (2) a cross-modal mapping block to enable a reliable complement of missing modality features.
\item For holistic representation learning in 3D genome analysis, we present \textbf{\textit{the largest paired dataset of Hi-C and epigenomic tracks}}, comprising over 1 million samples.
\item Extensive experiments demonstrate that MIX-HIC achieves state-of-the-art performance on three critical downstream tasks across two cell lines.
\end{itemize}

\vspace{-4mm}
\section{Related Works}
\vspace{-2mm}
\subsection{3D Genome-Related Tasks}
This work evaluates the effectiveness of MIX-HIC on three downstream tasks, including Hi-C contact map prediction, chromatin loop detection, and CAGE-seq expression prediction.

Existing methods for Hi-C contact map prediction from DNA sequences show promise but lack cross-cell-type generalization \cite{fudenberg2020predicting, schwessinger2020deepc}. Models like EPCOT \cite{zhang2023generalizable} and C.Origami \cite{tan2023cell} improve this by integrating DNA sequence with cell-type-specific epigenomic tracks. Epiphany \cite{yang2023epiphany} offers a more efficient solution using only the epigenomic tracks.

Chromatin loop detection methods are broadly divided into statistical and supervised learning methods. Statistical methods like HiCExplorer \cite{wolff2020galaxy} and ChromoSight \cite{matthey2020computer} rely on contact frequency distributions or expert-defined templates to identify loops. Supervised learning methods, including Peakachu \cite{salameh2020supervised}, DLoopCaller \cite{wang2022dloopcaller} leverage labeled data and sophisticated architectures for loop detection. 
RefHiC \cite{zhang2023reference} employs a coarse-to-fine training strategy that integrates multi-resolution Hi-C contact maps through contrastive learning mechanisms for model pre-training.
However, RefHiC is limited by its reliance on small-scale data semantics, which hinders its generalizability to other downstream tasks. 

CAGE-seq expression prediction typically uses multimodal inputs, including DNA sequences, epigenomic tracks, and Hi-C contact maps. Enformer \cite{avsec2021effective} employs transformers for modeling DNA sequences, while EPCOT \cite{zhang2023generalizable}  integrates DNA sequences and epigenomic tracks with transformers or long short-term memory (LSTM) \cite{hochreiter1997long}. GraphReg \cite{karbalayghareh2022chromatin} integrates epigenomic tracks and Hi-C contact maps using graph attention networks for expression prediction.

Despite their notable achievements, these methods often exhibit limitations in knowledge transfer across tasks and fail to fully capture the complex interaction patterns between multimodal data.

\vspace{-2mm}
\subsection{Foundation Models in Computational Biology}
\vspace{-2mm}
The emergence of foundation models has significantly advanced various fields of computational biology. For example, EvoRank \cite{zhuang2024pre} has demonstrated remarkable capabilities by harnessing extensive protein sequence datasets to derive latent representations through sequence alignment. VQDNA \cite{icml2024vqdna} employs large-scale DNA sequences to learn adaptive tokenization through vector-quantized codebooks. To capture more comprehensive views of data, multimodal foundation models such as ESM-IF \cite{hsu2022learning} combine protein sequences and 3D structures to learn functional and structural representations, advancing protein function design and prediction. Similarly, UniCorn \cite{feng2024unicorn} integrates 2D and 3D molecular views through contrastive learning, effectively capturing complementary bimodal features. These foundation models have showcased strong representation learning capabilities of foundation models, inspiring the development of a universal model capable of effectively addressing various downstream tasks in the 3D genomic field. 

\begin{wraptable}{r}{0.5\textwidth}
\vspace{-6.5mm}
\centering
\caption{Summary of pre-training data for MIX-HIC, including original and cleaned data counts.}
\label{tab:num_pretrain}
\vskip 0.1in
\small
\begin{tabular}{c|c|c|c}
\toprule
Cell line & Original    & Cleaned     & Remain \\ 
\midrule
HepG2     & 48, 415     & 38, 536     & 79.5\% \\
HCT116    & 276, 384    & 189, 099    & 68.4\% \\
IMR90     & 471, 557    & 316, 794    & 67.2\% \\
WTC11     & 1, 032, 048 & 731, 519    & 70.9\% \\ \midrule
\rowcolor{blue!10}
Total     & \textbf{1, 828, 401} & \textbf{1, 275, 948} & 69.8\% \\ \bottomrule
\end{tabular}
\vspace{-1.5mm}
\end{wraptable}

\vspace{-3mm}
\section{Data Processing and Preparation}
\vspace{-3mm}
To distill the comprehensive semantics from the 3D genome, we collect and refine a large-scale dataset for pre-training MIX-HIC, using publicly available data from the hg38 assembly. The Hi-C contact maps are obtained from the 4DN Data Portal\footnote{\url{https://data.4dnucleome.org/}}, while the epigenomic tracks (ATAC-seq and DNase-seq, which measure how `open' or accessible DNA is for transcription), CAGE-seq expression data (which directly quantifies gene activity levels), and CTCF ChIA-PET \cite{li2010chia} chromatin loops (which identify high-confidence interactions mediated by the key architectural protein CTCF) are downloaded from the ENCODE Portal \footnote{\url{https://www.encodeproject.org/}}. 
Due to the high cost of deep sequencing for Hi-C experiments \cite{chang2024droplet}, publicly available datasets are typically limited to resolutions of 5 kb, 10 kb, or coarser. A 5 kb resolution is a fine-grained and effective choice for deep learning models \cite{zhang2023generalizable, karbalayghareh2022chromatin}.
The MIX-HIC model processes 250,000 base pair (bp) genomic windows, a size selected to encompass key regulatory structures like chromatin loops, ensuring most functional units are fully contained within the inputs \cite{yang2023epiphany, salameh2020supervised}. This results in Hi-C contact maps being represented as $50 \times 50$ matrices at the 5 kb resolution. To reduce data variability, epigenomic tracks are averaged over every 100 bps, which generates 2,500-length sequences.
Each bin in the Hi-C matrix corresponds to specific $x$ and $y$ coordinates, representing two distinct genomic segments. Their respective epigenomic sequences are concatenated into a 5,000-length representation for that bin. Thus, Hi-C contact maps and epigenomic tracks are treated as images and sequences, respectively. We focus on four cell lines for pre-training, including HepG2, HCT116, IMR90, and WTC111. To ensure data quality, we filter out windows with fewer than 10\% non-zero Hi-C interactions or insufficient contact signals, which is a standard quality control step to remove uninformative data \cite{salameh2020supervised, wang2022dloopcaller}. Hi-C contact maps are inherently sparse, especially for long-range interactions distant from the diagonal. Including these extremely sparse, low-signal windows would introduce noise and degrade the model's training process. As summarized in Table~\ref{tab:num_pretrain}, this filtering process removed approximately 30\% of the raw windows. Crucially, a massive and high-quality dataset of over 1.2 million sample pairs was retained, which is more than sufficient for robust pre-training.
\textit{Further details on data processing, reference number, and downstream task data are provided in Appendix~\ref{sec:data_source}.}

\begin{figure*}[t]
\begin{center}
\centerline{\includegraphics[width=0.75\linewidth]{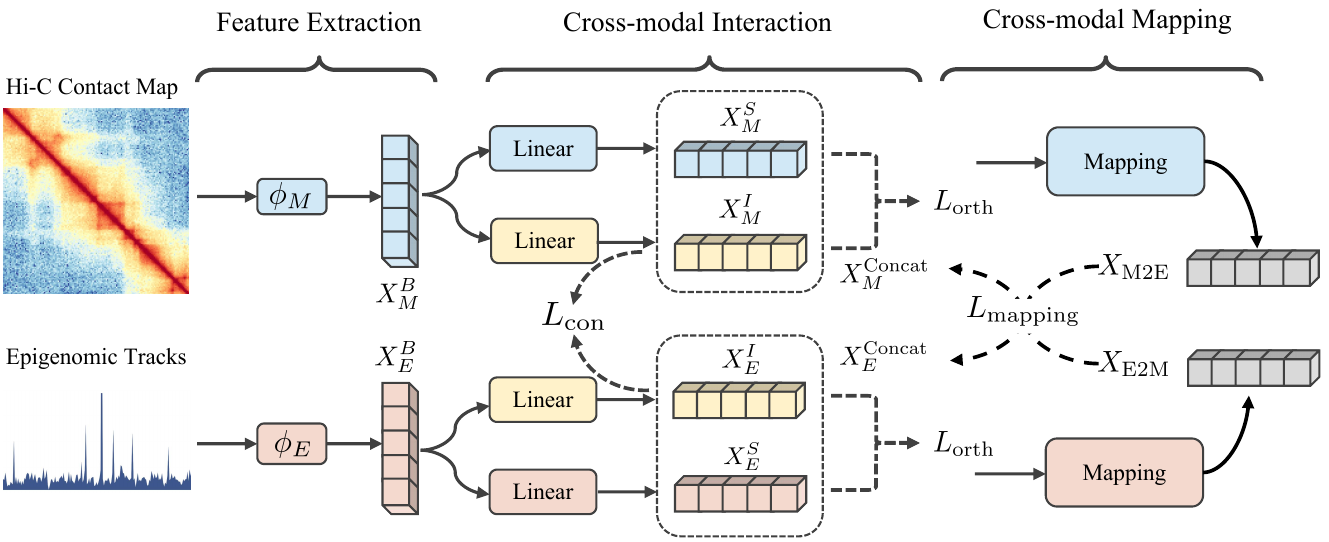}}
\vspace{-1mm}
\caption{\textbf{Pre-training stage of MIX-HIC.} MIX-HIC employs a dual-encoder architecture to extract refined features from both Hi-C contact maps and epigenomic tracks. The modal-specific and modal-invariant representations are learned via contrastive learning and orthogonal constraints within the cross-modal interaction block. A cross-modal mapping block is developed to further regularize the bimodal representations and facilitate cross-modal complement. $\phi_M$ and $\phi_E$ denote the Hi-C contact map encoder and epigenomic track encoder, respectively.}
\label{pretrain}
\end{center}
\vspace{-5.3mm}
\end{figure*}

\vspace{-3mm}
\section{Methodology}
\vspace{-3mm}
\subsection{Self-supervised Pre-training}
\vspace{-2mm}
Self-supervised learning \cite{simclr, wang2020enaet, wang2022importance} offers a powerful solution for learning unified and comprehensive representations from heterogeneous 3D genomic and epigenomic data. By leveraging large-scale pairwise data during pre-training, MIX-HIC effectively captures inherent biological patterns and relationships across these bimodal data, significantly enhancing its adaptability to diverse downstream tasks. As shown in Figure~\ref{pretrain}, the pre-training phase includes a feature extraction block, cross-modal interaction block, and cross-modal mapping block.
The feature extraction block employs specialized encoders for epigenomic tracks and Hi-C contact maps, respectively, to generate refined representations. The learned representations are then fed into the cross-modal interaction block to learn both modal-specific and modal-invariant features. This learning process is regularized through a combination of contrastive learning loss and orthogonal loss. Finally, a cross-modal mapping block further explores latent connections and complementary information between the two modalities.

\textbf{Feature extraction block.}
The feature extraction block consists of Hi-C and epigenomic feature encoders. Given the Transformer's capability to model long-range dependencies and its flexibility in capturing both spatial interactions (Hi-C contact maps) and sequential relationships (epigenomic tracks), we utilize a Transformer-based architecture for robust feature extraction and multimodal integration. Similar to Vision Transformer (ViT) \cite{dosovitskiy2020image}, the single-channel Hi-C contact map $X_{M} \in \mathbb{R}^{1 \times H \times W}$ is first transformed into a sequence of flattened patches $X_{M}^{p} \in \mathbb{R}^{N \times D}$. Here, $H$ and $W$ are the height and width of the Hi-C contact map, both equal to 50. The dimension $D=1\times P \times P$ corresponds to the initial size of each patch, where the patch size $P$ is set to 2. The total number of these patches $N=H \times W / P^2$ is the resulting number of patches, which becomes the input sequence length. In the Hi-C feature encoder, a feedforward network projects these patches into an embedding $X_{M}^{0} \in \mathbb{R}^{N \times C}$, where $C$ denotes the predefined feature dimension. This embedding is then refined through three cascaded encoder layers, where each layer consists of $T$ Transformer blocks followed by a downsampling layer, yielding progressively refined embeddings  $X_{M}^{i} \in \mathbb{R}^{\alpha_{i} \times C_i}$, where the sequence length $\alpha_{i}$ is defined as $\alpha_{i} = \frac{N}{4^i}$ and the feature dimension $C_i$ grows exponentially as $C_i = 2^{i}C$, with the encoder layer $i \in \{1,2,3\}$. Finally, a bottleneck layer equipped with $T$ Transformer blocks is applied to produce the Hi-C contact map embedding $X_{M}^{B} \in \mathbb{R}^{\alpha_{3} \times C_3}$.

The epigenomic encoder processes input sequences $X_{E} \in \mathbb{R}^{L_1 \times O}$, where $L_1=5,000$ is the initial sequence length, and $O$ represents the two epigenomic tracks: ATAC-seq and DNase-seq. Convolutional layers with max-pooling operations extract an initial embedding $X_{E}^{0} \in \mathbb{R}^{L_2 \times C}$, where $L_2=100$ corresponds to two genomic segments along both $x$ and $y$ axes in the Hi-C contact map. Note that processing epigenomic tracks at high resolution (\textit{e.g.}, 100bp) is common practice to avoid over-smoothing \cite{yang2023epiphany, zhang2023generalizable}. Similarly, the embedding $X_{E}^{0}$ is processed  through three encoder layers, where each layer consists of $T$ Transformer blocks followed by a downsampling layer, resulting in
$X_{E}^{i} \in \mathbb{R}^{\beta_{i} \times C_i}$, where the sequence length $\beta_{i} = \frac{L_2}{2^i}$ for $i \in \{1,2,3\}$. Then, the refined epigenomic representation $X_{E}^{\text{B}} \in \mathbb{R}^{\beta_3 \times C_3}$ is also derived from a final bottleneck layer.

\textbf{Cross-modal interaction block.}
Conventional multimodal learning architectures \cite{yang2024employing} often utilize contrastive learning to project features from different modalities into a shared embedding space. However, these modalities inherently contain both homogeneous and heterogeneous information. Direct aligning all features risks losing essential modal-specific characteristics, potentially diminishing downstream task performance \cite{dong2023simmmdg, wang2023multi}. This is further analyzed in Theorem~1 as follows.

\textbf{Theorem 1.} \textit{Let \( \phi_1 \) and \( \phi_2 \) be feature encoders for two modalities \( z_1\) and \(z_2 \), respectively. If the encoded features \( \mathbf{F}^1 = \phi_1(z_1) \) and \( \mathbf{F}^2 = \phi_2(z_2) \) are perfectly aligned such that \( \mathbf{F}^1 = \mathbf{F}^2 \), we have:}

\vspace{-2mm}
\begin{equation}
    \inf_h \mathbb{E}_q[\mathcal{L}_{\text{CE}}(h(\mathbf{F}^1, \mathbf{F}^2), t)] - \inf_{h'} \mathbb{E}_q[\mathcal{L}_{\text{CE}}(h'(z_1, z_2), t)] \geq \Gamma_q.
\label{equation：theorem}
\end{equation}
\vspace{-2mm}

\textbf{Remarks.}
The information gap \( \Gamma_q := \max\{U(z_1; t), U(z_2; t)\} - \min\{U(z_1; t), U(z_2; t)\} \) quantifies the effectiveness of the modalities in predicting target variable \( t \), where \( U(z_j; t) \) represents the mutual information. Here, \( \mathbb{E} \) represents the expectation, \( q \) denotes the joint distribution of \( (z_1, z_2, t) \), \( \mathcal{L}_{\text{CE}} \) is the cross-entropy loss, and \( h \) and \( h' \) are prediction functions for features and raw data, respectively. Theorem 1 demonstrates that perfect alignment results in prediction errors that are suboptimal by at least \( \Gamma_q \) compared to using raw modalities directly. This information gap widens when information content is imbalanced across modalities. The complete proof is provided in Appendix~\ref{sec:Analysis}.

To address this, our cross-modal interaction block captures both modal-specific and modal-invariant representations, enabling a more comprehensive understanding of the data. Specifically, we employ four independent dense networks to process the Hi-C contact map representation $X_{M}^{B}$ and the epigenomic representation $X_{E}^{B}$. This generates the condensed modal-invariant representations $X_{M}^{I} \in \mathbb{R}^{\alpha_3 \times C_2}$ and $X_{E}^{I} \in \mathbb{R}^{\beta_3 \times C_2}$, as well as the modal-specific representations $X_{M}^{S} \in \mathbb{R}^{\alpha_3 \times C_2}$ and $X_{E}^{S} \in \mathbb{R}^{\beta_3 \times C_2}$. The mean pooling operation is then applied along the sequence length dimension, producing $\hat{X_{M}^{I}}$, $\hat{X_{E}^{I}}$, $\hat{X_{M}^{S}}$, and $\hat{X_{E}^{S}}$, each with a feature dimension of $C_2$. To ensure that the modal-invariant features capture shared knowledge across modalities, we incorporate a contrastive learning loss to regularize these representations. We propose a unified contrastive loss function $\mathcal{L}_{\text{pair}}$ \cite{simclr} to compute the similarity between two modalities $\mathcal{A}$ and $\mathcal{B}$ as follows:
\vspace{-1mm}
\begin{equation}
\mathcal{L}_{\text{pair}}(\mathcal{A}, \mathcal{B}) = -\frac{1}{J} \sum_{j=1}^{J} \text{log} \frac{\text{exp} \langle \mathcal{A}_j, \mathcal{B}_j \rangle / \tau}{\sum_{r=1}^{J} \text{exp} \langle \mathcal{A}_j, \mathcal{B}_r  \rangle / \tau},
\end{equation}
\vspace{-1mm}


where $\tau$ denotes the temperature, commonly set to 0.07 \cite{kim2023neural}, $J$ refers to the batch size,$ \langle \cdot, \cdot \rangle $ is the dot product operation, and $\mathcal{A}_j$ and $\mathcal{B}_j$ represent the embedding of the $j$-th sample in the mini-batch for modality $\mathcal{A}$ and $\mathcal{B}$, respectively. Finally, the overall contrastive loss is computed as follows:
\begin{equation}
\mathcal{L}_{\text{con}} = \frac{1}{2}(\mathcal{L}_{\text{pair}}(\hat{X_{E}^{I}},\hat{X_{M}^{I}}) + \mathcal{L}_{\text{pair}}(\hat{X_{M}^{I}},\hat{X_{E}^{I}})).
\end{equation}

Moreover, we introduce an orthogonal constraint to maximize the dissimilarity between modal-specific and modal-invariant features. This ensures that the modal-specific features capture complementary information distinct from the shared knowledge represented by the modal-invariant features. The orthogonal loss $\mathcal{L}_{\text{orth}}$ is calculated by minimizing the inner product between these features:
\begin{equation}
\mathcal{L}_{\text{orth}} = \frac{1}{2} \left( \langle \hat{X_{M}^{S}}, \hat{X_{M}^{I}} \rangle + \langle \hat{X_{E}^{S}}, \hat{X_{E}^{I}} \rangle \right).
\end{equation}

\textbf{Cross-modal mapping block.}
Data scarcity, often due to high experimental costs, can lead to incomplete datasets with missing modalities. Integrating predicted features of a missing modality with existing modalities can enhance prediction performance \cite{wang2023multi, kim2025missing}. Our cross-modal mapping block aims to capture the implicit semantic relationships and facilitate knowledge transfer between modalities to address this issue. 

Some downstream tasks require the MIX-HIC to preserve the sequence length of the input features. However, the differing lengths of Hi-C contact maps and epigenomic tracks pose a challenge for effective modality transfer. Therefore, we apply 1D adaptive pooling to the concatenated representations of Hi-C, $ X_{M}^{\text{Concat}} = [X_{M}^{I}:X_{M}^{S}] \in \mathbb{R}^{\alpha_3 \times C_3} $ and epigenomic tracks, $ X_{E}^{\text{Concat}} = [X_{E}^{I}:X_{E}^{S}] \in \mathbb{R}^{\beta_3 \times C_3} $ (where $[\cdot:\cdot]$ denotes concatenation), to align their lengths. This yields the complementary representations $X_{\text{M2E}} \in \mathbb{R}^{\beta_3 \times C_3}$ and $X_{\text{E2M}} \in \mathbb{R}^{\alpha_3 \times C_3}$ as follows:
\begin{equation}
X_{\text{M2E}} = \mathcal{F}_{\text{M2E}}(\mathcal{G}_{\text{M2E}}(X_{M}^{\text{Concat}})), X_{\text{E2M}} = \mathcal{F}_{\text{E2M}}(\mathcal{G}_{\text{E2M}}(X_{E}^{\text{Concat}})),
\label{e2m}
\end{equation}
where $\mathcal{G}_{\text{M2E}}$ and $\mathcal{G}_{\text{E2M}}$ are 1D adaptive pooling operations. $\mathcal{F}_{\text{M2E}}$ and $\mathcal{F}_{\text{E2M}}$ denote dense layers with the same output dimensions matching their inputs.
To ensure these mapped embeddings capture the relevant information from the target modality, we use a
$\mathcal{L}_{\text{mapping}}$ loss for regularization:
\vspace{-1mm}
\begin{equation}
\mathcal{L}_{\text{mapping}} = \frac{1}{2}(\left\| X_{\text{M2E}} - X_{E}^{\text{Concat}} \right\|_2^2 + \left\| X_{\text{E2M}} - X_{M}^{\text{Concat}} \right\|_2^2).
\end{equation}

\vspace{-2mm}
Overall, the final loss for self-supervised pre-training is computed as follows:
\begin{equation}
\mathcal{L}_{\text{pretrain}} = \mathcal{L}_{\text{con}} + \mathcal{L}_{\text{orth}} + \mathcal{L}_{\text{mapping}}.
\end{equation}

\subsection{Task-specific Fine-tuning}
\begin{wrapfigure}{r}{0.495\textwidth}
\vspace{-3.5mm}
\begin{center}
\centerline{\includegraphics[width=1.0\linewidth]{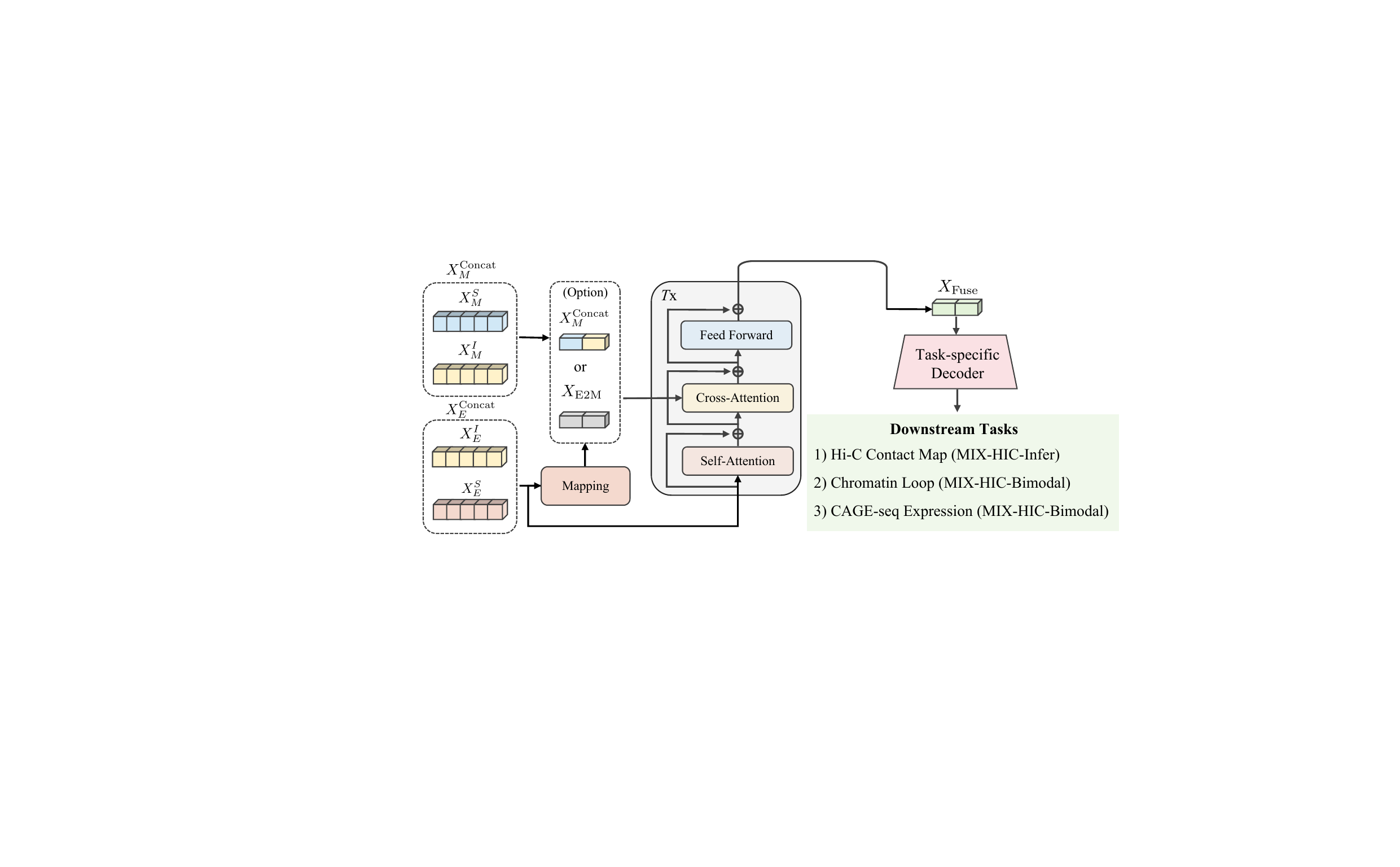}}
\caption{\textbf{Fine-tuning stage of MIX-HIC.} Epigenomic features are captured from the pre-trained encoder, while Hi-C contact map features are obtained either directly from the pre-trained encoder or through feature mapping based on the epigenomic features. MIX-HIC incorporates a modality fusion block to integrate the bimodal representations, followed by a task-specific decoder for final predictions of downstream tasks.}
\label{finetune}
\end{center}
\vspace{-10mm}
\end{wrapfigure}

As depicted in Figure~\ref{finetune}, MIX-HIC is a versatile framework capable of processing various kinds of inputs. Ideally, MIX-HIC takes both the Hi-C contact map and the epigenomic feature profile as inputs (represented as MIX-HIC-Bimodal), extracting their concatenated features $X_{M}^{\text{Concat}}$ and $X_{E}^{\text{Concat}}$ similar to the feature extraction block used in the self-supervised pre-training stage. However, in real-world scenarios, a certain modality may be absent for various unforeseen reasons. 

Leveraging the powerful representation ability of pre-training to capture implicit connections between bimodal data, MIX-HIC incorporates a cross-modal mapping block to complement the features of the missing modality using the information from the available modality. For example, when the Hi-C contact map is missing, MIX-HIC can infer the missing modality features $X_{\text{E2M}}$ from the concatenated epigenomic embedding $X_{E}^{\text{Concat}}$ using Eq. \ref{e2m}. For the Hi-C contact map prediction task, even if only the epigenomic tracks are available, the corresponding contact map features $X_{\text{E2M}}$ can still be utilized for prediction (denoted as MIX-HIC-Infer).

\textbf{Modality-fusion block.} We employ $T$ stacked contact map-grounded fusion blocks to learn the interaction patterns between the bimodal representations. Each contact map-grounded fusion block consists of a self-attention layer, a cross-attention layer, and a feedforward network. The extracted epigenomic embeddings $X_{E}^{\text{Concat}}$ are first input into the self-attention layer, generating query embeddings of the cross-attention layer. The contact map embeddings, either $X_{M}^{\text{Concat}}$ or $X_{\text{E2M}}$, serve as the key and value embeddings, which are then processed through the cross-attention layer. Finally, a feedforward neural network is applied to produce the fusion embeddings $X_{\text{Fuse}} \in \mathbb{R}^{\beta_3 \times C_3}$.

\textbf{Task-specific prediction block.} Three types of decoders are involved in the task-specific prediction block. The chromatin loop detection task is a binary classification problem. Therefore, the decoder for this task includes a mean pooling operation along the sequence length axis, followed by a feedforward network, to classify a given sample as either a chromatin loop or a non-loop.

The CAGE-seq expression task is formulated as a regression problem, aiming to predict 100 values corresponding to the expression levels of two genomic segments (each of length 50) along $x$-axis and $y$-axis of the Hi-C contact map. The decoder consists of three transformer blocks, with each followed by an upsampling layer. At each stage, the encoder-derived features $ X_E^{i} $ from the $ i $-th encoder layer are concatenated with the corresponding $ i $-th layer of decoder outputs $ X_D^{i} $ from the preceding stage via skip connections, facilitating feature integration across network depths. Finally, the decoder output features $ X_{\text{out}} \in \mathbb{R}^{L_2 \times 2C} $ are processed through two feedforward networks to generate regression predictions.

To predict the $50 \times 50$ Hi-C contact maps at 5kb resolution, the decoder output features $ X_{\text{out}} $ are obtained similar to the CAGE-seq expression task. These features are then split into two feature segments $ X_{\text{out}}^1 \in \mathbb{R}^{50 \times 2C}$ and $ X_{\text{out}}^2 \in \mathbb{R}^{50 \times 2C}$ along the length axis. $ X_{\text{out}}^1 $ and $ X_{\text{out}}^2 $ are unsqueezed into $ \mathbb{R}^{50 \times 1 \times 2C} $ and $ \mathbb{R}^{1 \times 50 \times 2C} $, respectively, after which element-wise addition and multiplication operations are applied to generate feature maps of size $ \mathbb{R}^{50 \times 50 \times 2C} $. Finally, two layers of feedforward networks are utilized to predict the Hi-C contact maps.

\textbf{Training loss.} The chromatin loop detection task employs the binary cross entropy (BCE) loss function, while the Hi-C contact map prediction and CAGE-seq expression prediction tasks use the mean squared error (MSE) loss function. 
It should be noted that we normalize CAGE-seq data using RPGC \cite{ramirez2016deeptools2} and Hi-C data using KR \cite{kaul2020identifying} normalization to correct for sequencing depth and systematic biases, respectively. Subsequently, a log transformation is applied to both datasets. This is critical for stabilizing variance and compressing the highly skewed distribution of raw counts, ultimately producing continuous values representing normalized interaction frequencies.
Consequently, the MSE loss becomes an appropriate and robust choice for these continuous values. The MSE and BSE loss functions can be formulated as follows:
\vspace{-1mm}
\begin{equation}
\mathcal{L}_{\text{MSE}} = \frac{1}{J} \sum_{j=1}^{J} (y_j - \hat{y}_j)^2,
\end{equation}
\vspace{-3mm}
\begin{equation}
\mathcal{L}_{\text{BCE}} = - \frac{1}{J} \sum_{j=1}^{J} \left[ y_j \log(\hat{y}_j) + (1 - y_j) \log(1 - \hat{y}_j) \right],
\end{equation}

\vspace{-3mm}
where $y_i$ and $ \hat{y}_i $ represent the true value and the predicted value, respectively.

\vspace{-3mm}
\section{Experiments}
\vspace{-2.5mm}
\textit{Implementation details, hyperparameter analysis, and analyses of the model's biological grounding and robustness for noise can be referred to Appendix.}
\vspace{-3.5mm}

\subsection{Comparison Results on Downstream Tasks}
\vspace{-2mm}
To demonstrate the effectiveness of MIX-HIC, three downstream tasks are involved in this work, including Hi-C contact map prediction, chromatin loop detection, and CAGE-seq expression prediction. We construct three versions of MIX-HIC: (1) MIX-HIC-Bimodal, which leverages both Hi-C contact maps and epigenomic tracks through pre-training; (2) MIX-HIC-NonPre, a non-pretrained version using the same bimodal inputs; and (3) MIX-HIC-Infer, designed to handle missing Hi-C data by integrating epigenomic track embeddings with inferred Hi-C embeddings. \textit{Detailed descriptions of the compared state-of-the-art methods are given in Appendix~\ref{sec:assets}.}

\begin{wraptable}{r}{0.46\textwidth}
\vspace{-6.5mm}
\centering
\caption{Methods comparison for the Hi-C contact map prediction task on GM12878 and K562 cell lines using $R^2$. The results marked in \textbf{bold} and \underline{underlined} denote the best and second-best performing methods, respectively.}
\label{tab:contact}
\vskip 0.05in
\small
\renewcommand{\arraystretch}{1.2}
\resizebox{\linewidth}{!}{
\begin{tabular}{l|l|l}
\toprule
Methods                                              & GM12878                         & K562                                \\ \midrule
Epiphany \cite{yang2023epiphany}                                            & 0.7970                          & 0.6547                              \\
C.Origami \cite{tan2023cell}                                           & 0.7958                          & 0.7055                              \\
EPCOT-Transformer \cite{zhang2023generalizable}                                   & 0.5409                          & 0.7648                              \\
EPCOT-LSTM \cite{zhang2023generalizable}                                          & \underline{0.7993} & \underline{0.7840} \\
\rowcolor{blue!10} MIX-HIC-Infer (Ours) & $\textbf{0.8724}_{\textcolor{red}{+9.3\%}}$                 &  ${\textbf{0.8001}}_{\textcolor{red}{+2.1\%}}$ \\ \bottomrule
\end{tabular}}
\vspace{-4mm}
\end{wraptable}

\textbf{3D Chromatin Organization Prediction.} In this task, the Hi-C contact maps serve as the target. MIX-HIC-Infer is compared with four state-of-the-art methods, including Epiphany \cite{yang2023epiphany}, C.Origami \cite{tan2023cell}, and two variants of EPCOT \cite{zhang2023generalizable}, (\textit{i.e.} EPCOT-LSTM and EPCOT-Transformer). For a fair comparison, all methods receive the same input as MIX-HIC and are configured with their default parameter settings. 
The coefficient of determination ($R^2$) \cite{nagelkerke1991note} is employed as the evaluation metric, since it effectively quantifies explained variance in analyses of long-tailed or sparse data, such as low-probability long-range interactions in Hi-C.

Table~\ref{tab:contact} shows the evaluation performance on GM12878 and K562 cell lines. Compared to other methods, MIX-HIC-Infer demonstrates superior performance, achieving the highest average $R^2$ values on both GM12878 and K562 cell lines. 
Specifically, it outperforms the runner-up method by approximately 9.3\% and 2.1\% in $R^2$ score on GM12878 and K562, respectively. 
We note that the EPCOT variants deliver competitive performance, but they suffer from fluctuations across the two datasets. Through extensive pre-training on large-scale pairwise datasets, MIX-HIC effectively explores implicit semantic relationships between bimodal data, enabling robust compensation for missing modality semantics and achieving promising accuracy in Hi-C contact map prediction.

\textbf{Chromatin Loop Detection.} Two supervised machine learning-based methods DLoopCaller \cite{wang2022dloopcaller} and Peakachu \cite{salameh2020supervised} are used to compare with MIX-HIC-Bimodal. As illustrated in Table~\ref{tab:super-loops}, MIX-HIC-Bimodal surpasses other machine learning-based methods across all classification metrics on two datasets. By integrating both epigenomic tracks and Hi-C contact maps, DLoopCaller slightly outperforms Peakachu, which relies solely on Hi-C contact maps. The enhanced performance of MIX-HIC can be attributed to the effective self-supervised pre-training, which facilitates the capture of both modal-invariant and modal-specific information from bimodal representations, in contrast to the simple concatenation employed by DLoopCaller.

\begin{wrapfigure}{r}{0.45\textwidth}
\vspace{-8mm}
\begin{center}
\centerline{\includegraphics[width=1.0\linewidth]{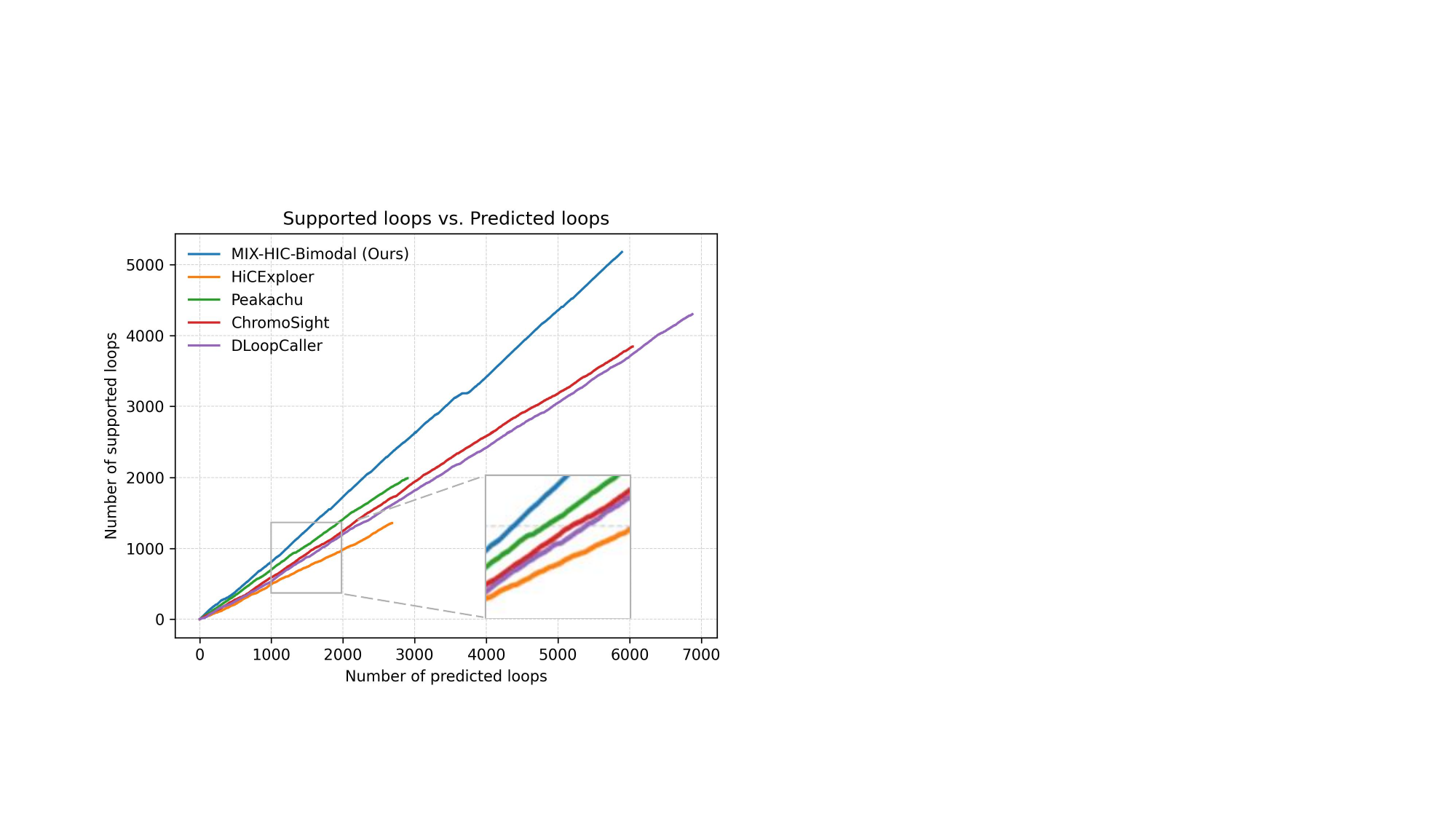}}
\vspace{-2mm}
\caption{Comparison of the number of predicted loops with the number of corresponding ChIA-PET-supported loops across various deep learning methods on GM12878 cell line.}
\label{fig:loop_GM12878}
\end{center}
\vspace{-5mm}
\end{wrapfigure}

We employ both statistical-based methods, namely ChromoSight \cite{matthey2020computer} and HiCExplorer \cite{wolff2020galaxy}, as well as machine learning-based methods to annotate loops across entire chromosomes. The predicted loops are further compared with those loops validated by experimental data from ChIA-PET. \textit{Method details for whole-chromosome chromatin loop annotation and corresponding results with the quantitative `proportion' metric are available in Appendix \ref{sec:whole_chrom} and \ref{sec:proportion}, respectively}. Figure~\ref{fig:loop_GM12878} compares the number of predicted loops versus the number of loops supported by ChIA-PET among various methods on the GM12878 dataset. Among all the methods, machine learning-based methods generally predict a higher number of loops that are validated by ChIA-PET on the GM12878 dataset, with MIX-HIC identifying the most ChIA-PET validated loops while maintaining the highest validated proportion. 
On the K562 dataset, which contains much less training data compared to GM12878, the efficacy of machine learning methods generally declines. Nevertheless, MIX-HIC consistently maintains the highest proportion of validated loops in comparison with other methods. 
\textit{An example of whole-chromosome chromatin loop comparison is provided in Appendix \ref{sec:example_loop}.}

\begin{table*}[t]
\centering
\caption{Methods comparison for the supervised chromatin loop detection task on GM12878 and K562 cell lines.}
\label{tab:super-loops}
\vskip 0.1in
\renewcommand{\arraystretch}{1.2}
\begin{footnotesize}
\resizebox{\textwidth}{!}{
\begin{tabular}{l|llll|llll}
\toprule
                         & \multicolumn{4}{c|}{GM12878}                                          & \multicolumn{4}{c}{K562}                                                                                                                                        \\ \cmidrule{2-9} 
\multirow{-2}{*}{Methods} & \multicolumn{1}{l}{Precision}       & \multicolumn{1}{l}{Recall}          & \multicolumn{1}{l}{F1}              & \multicolumn{1}{l|}{AUROC}           & \multicolumn{1}{l}{Precision}                            & \multicolumn{1}{l}{Recall}                               & \multicolumn{1}{l}{F1}                                      & \multicolumn{1}{l}{AUROC}                                   \\ \midrule
Peakachu \cite{salameh2020supervised}                & 0.7763          & \underline{0.8283}    & 0.8015          & 0.8766          & 0.7895                               & \underline{0.7905}                      & 0.7900                            & 0.8834                                  \\
DLoopCaller \cite{wang2022dloopcaller}             & \underline{0.8433} & 0.8075          & \underline{0.8250}    & \underline{0.9046}    & \underline{0.8383} & 0.7526       & \underline{0.7932}          & \underline{0.8924}    \\
\rowcolor{blue!10}
MIX-HIC-Bimodal (Ours)          & $\textbf{0.8505}_{\textcolor{red}{+0.9\%}}$    & $\textbf{0.8337}_{\textcolor{red}{+0.7\%}}$ & $\textbf{0.8420}_{\textcolor{red}{+2.1\%}}$ & $\textbf{0.9209}_{\textcolor{red}{+1.8\%}}$ & $\textbf{0.8521}_{\textcolor{red}{+1.6\%}}$                      & $\textbf{0.8027}_{\textcolor{red}{+1.5\%}}$ & $\textbf{0.8267}_{\textcolor{red}{+3.9\%}}$ & $\textbf{0.9194}_{\textcolor{red}{+3.0\%}}$ \\ \bottomrule
\end{tabular}}
\end{footnotesize}
\end{table*}

\begin{wraptable}{r}{0.448\textwidth}
\vspace{-6.5mm}
\centering
\caption{Methods comparison for the CAGE-seq expression prediction task on GM12878 and K562 cell lines using $R^2$.}
\label{tab:cage}
\vskip 0.1in
\renewcommand{\arraystretch}{1.2}
\small
\resizebox{\linewidth}{!}{
\begin{tabular}{l|l|l}
\toprule
Methods                                     & \multicolumn{1}{l|}{GM12878}    & \multicolumn{1}{l}{K562}        \\ \midrule
EPI-CNN \cite{karbalayghareh2022chromatin}                                    & 0.7719                          & 0.8033                          \\
EPI-Graph \cite{karbalayghareh2022chromatin}                                  & 0.7965                          & 0.8211                          \\
EPCOT-LSTM \cite{zhang2023generalizable}                                 & 0.4723                          & \underline{0.8704} \\
EPCOT-Transformer \cite{zhang2023generalizable}                          & \underline{0.8578} & 0.8230                          \\
\rowcolor{blue!10} MIX-HIC-Bimodal (Ours) & $\textbf{0.8833}_{\textcolor{red}{+3.0\%}}$                 & $\textbf{0.9077}_{\textcolor{red}{+4.3\%}}$                 \\ \bottomrule
\end{tabular}}
\end{wraptable}

\textbf{CAGE-seq Expression Prediction.} MIX-HIC-Bimodal is pitted against four benchmark methods: two variants of GraphReg (EPI-CNN and EPI-Graph) \cite{karbalayghareh2022chromatin}, as well as two variants of EPCOT (EPCOT-LSTM and EPCOT-Transformer) \cite{zhang2023generalizable}. EPI-CNN relies solely on epigenomic tracks to predict CAGE-seq expression, 
whereas EPI-Graph enhances this by incorporating Hi-C contact maps via graph attention networks for dual-modal modeling.
Table~\ref{tab:cage} showcases the comparison results across two datasets. EPCOT achieves competitive performance, yet it suffers from a time-consuming problem due to the process of long-range DNA sequence. MIX-HIC demonstrates remarkable performance over other benchmark methods across all metrics on both datasets. 

Overall, MIX-HIC outperforms state-of-the-art methods across all downstream tasks, with the most significant improvement (9.3\% $R^2$) in Hi-C contact map prediction. Large-scale pre-training enables robust semantic representations for MIX-HIC, while other methods lack generalizability due to being narrowly optimized for specific tasks and often show inconsistent performance across datasets.


\begin{wrapfigure}{r}{0.45\textwidth}
\vspace{-2mm}
\begin{center}
\centerline{\includegraphics[width=1.0\linewidth]{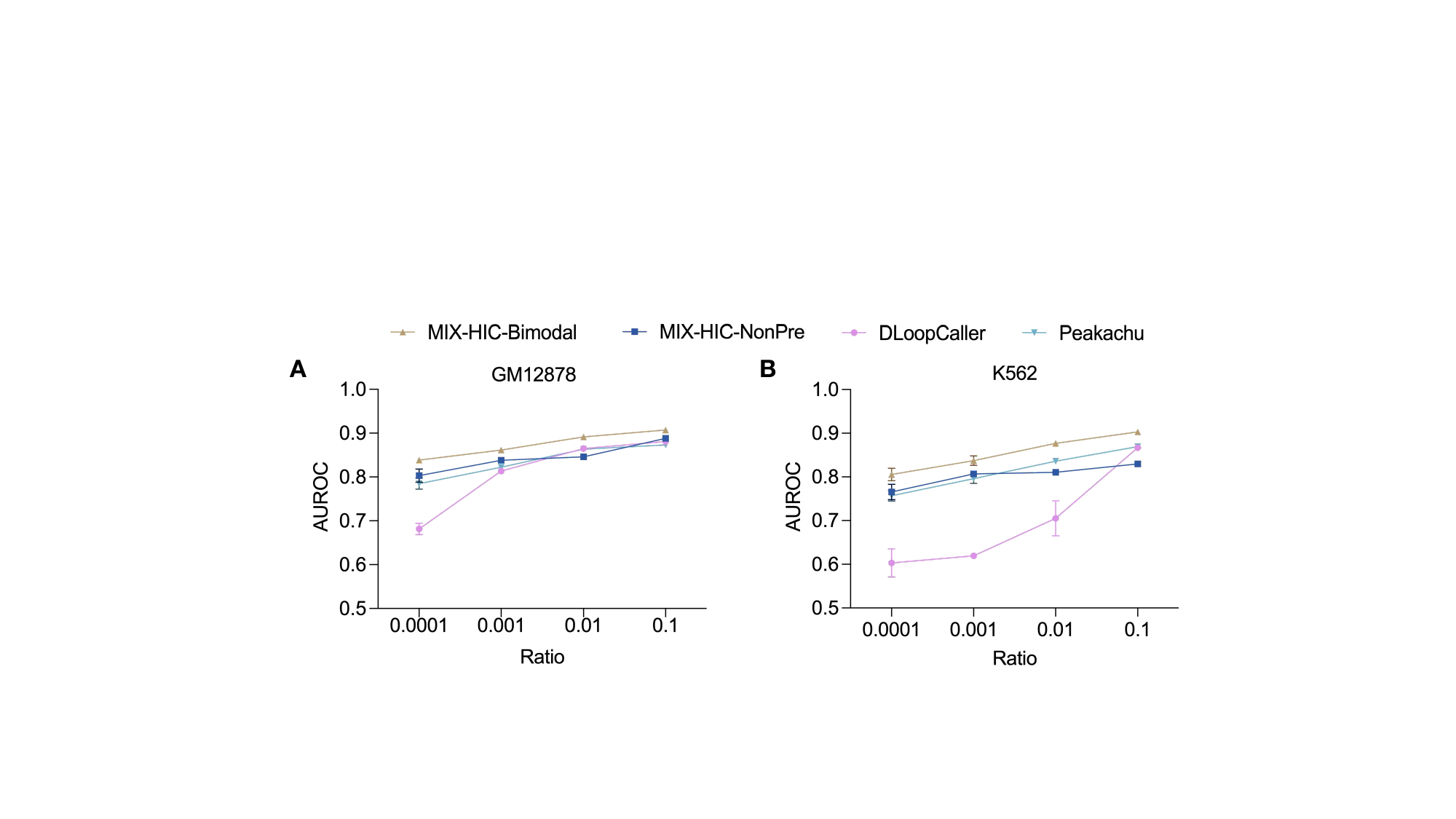}}
\vspace{-1mm}
\caption{Few-shot chromatin loop classification performance across different training data ratios. Mean values and standard errors are calculated over five independent runs with varying random seeds.}
\label{fig:fewshot}
\end{center}
\vspace{-9mm}
\end{wrapfigure}

\subsection{Few-shot Chromatin Loops Classification}

We conduct a few-shot learning experiment on the chromatin loops classification task to evaluate the performance of MIX-HIC under limited training data scenarios. Specifically, we fine-tune MIX-HIC-Bimodal, MIX-HIC-NonPre, DLoopCaller, and Peakachu using four different ratios of the training data, ranging from 0.0001 to 0.1. As illustrated in Figure~\ref{fig:fewshot}, MIX-HIC-Bimodal consistently outperforms other methods across all data ratios. Notably, the performance of most methods remains relatively robust even with minimal data, primarily stemming from the powerful and repetitive biological signatures of chromatin loops, which make the task tractable for most models. The performance variations among different architectures also highlight their inherent data efficiency. For instance, Peakachu, a random forest model, is inherently stable as it relies on engineered features that are less sensitive to data volume. In contrast, DLoopCaller's CNN architecture is more data-hungry and thus shows a steeper decline. Our MIX-HIC architecture, even without pre-training, proves more data-efficient: the self-attention mechanism is better suited for capturing the global and long-range dependencies in contact maps than local CNNs, and its bimodal input provides complementary information, enhancing robustness even in low-data regimes.

With a training data ratio of 0.1, MIX-HIC-Bimodal achieves an AUROC of about 0.9 on two datasets, which is competitive with other state-of-the-art methods trained on full datasets. These findings highlight the robustness and efficiency of MIX-HIC in leveraging pre-trained knowledge to achieve superior performance even with limited labels.

\begin{wrapfigure}{r}{0.45\textwidth}
\vspace{-6.5mm}
\begin{center}
\centerline{\includegraphics[width=1.0\linewidth]{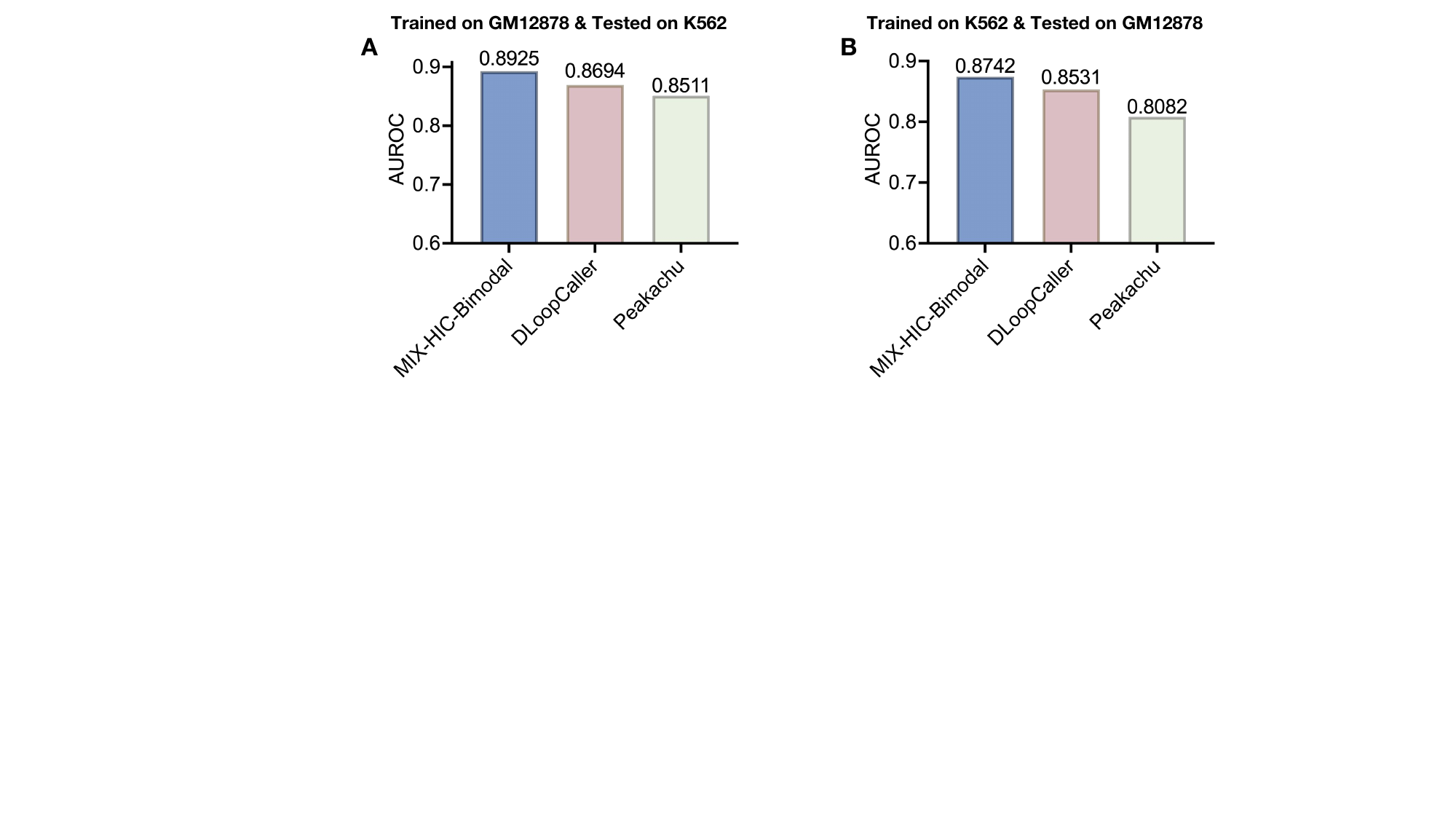}}
\vspace{-2.7mm}
\caption{Methods comparison for cross-cell-type evaluation.}
\label{fig:independent}
\end{center}
\vspace{-8mm}
\end{wrapfigure}

\subsection{Robust Performance Across Cell Types}
The utilization of cell-type specific data, \textit{i.e.}, Hi-C contact maps and epigenomic tracks, empowers MIX-HIC to achieve accurate predictions for novel cell types. We perform a cross-dataset evaluation with the GM12878 and K562 cell lines on the chromatin loop detection task to assess the generalization ability of MIX-HIC. In particular, all the models are trained on one cell line and evaluated on the other. The evaluation results of cross-cell type prediction are shown in Figure~\ref{fig:independent}. We note a decrease in performance when all models are evaluated on external datasets compared to within-dataset testing (see Table~\ref{tab:super-loops}). Nevertheless, MIX-HIC-Bimodal continues to outperform other methods, indicating its strong generalization ability in real-world scenarios.

\begin{wraptable}{r}{0.495\textwidth}
\vspace{-7.3mm}
\centering
\caption{AUROC results of ablation studies on loss terms for chromatin loop detection task. Symbols `\checkmark' and `-' denote present and absent, respectively.}
\label{tab:loss}
\vskip 0.1in
\renewcommand{\arraystretch}{1.2}
\resizebox{\linewidth}{!}{
\begin{tabular}{ccc|c|c}
\toprule
$\mathcal{L}_{\text{con}}$ & $\mathcal{L}_{\text{orth}}$ & $\mathcal{L}_{\text{mapping}}$ & GM12878  & K562  \\ \midrule
\checkmark        & -            & -           & 0.9136         & 0.9099 \\
\checkmark        & \checkmark   & -           & 0.9183         & 0.9156 \\
\checkmark        & \checkmark   & \checkmark  & \textbf{0.9209}         & \textbf{0.9194} \\ \bottomrule
\end{tabular}}
\vspace{-6mm}
\end{wraptable}

\subsection{Ablation Study}
To evaluate the effectiveness of key components in MIX-HIC, we perform ablation studies focusing on the proposed loss terms and the representation learning ability across multimodal data.

Three critical loss terms are employed during MIX-HIC pre-training (\textit{i.e.}, $\mathcal{L}_{\text{con}}$, $\mathcal{L}_{\text{orth}}$, and $\mathcal{L}_{\text{mapping}}$). We examine the effects of each loss component. As shown in Table~\ref{tab:loss}, the model equipped with $\mathcal{L}_{\text{con}}$ provides a strong baseline. The orthogonal loss enhances discrimination between modal-invariant and modal-specific representations, contributing about 0.5\% AUROC improvement compared to simply feature alignment, as demonstrated in Theorem~1. \textit{Further details for the orthogonal constaint are provided in Appendix~\ref{sec:inner}.} Although the cross-modal mapping loss provides modest enhancement, it enables missing modality inference of MIX-HIC. These findings demonstrate that robust multimodal fusion is achieved by explicitly separating modal-invariant and modal-specific representations using appropriate loss terms.

\begin{wraptable}{r}{0.495\textwidth}
\vspace{-6.3mm}
\centering
\caption{Ablation results of each modality, with values reported as $R^2$ (Hi-C contact map prediction), AUROC (chromatin loops dectection), and $R^2$ (CAGE-seq expression prediction). Symbol `$\circ$' represents inferred embeddings from the other modality.}
\label{tab:rep}
\vskip 0.1in
\renewcommand{\arraystretch}{1.2}
\resizebox{\linewidth}{!}{
\begin{tabular}{c|ccc|c|c}
\toprule
\multicolumn{1}{c|}{Tasks} & Epi.                      & Hi-C                      & Pre-trained               & GM12878                                                        & K562                                                           \\ \midrule
                         & \checkmark & -                         & -                         & 0.8481                                                         & 0.7709                                                         \\
\multirow{-2}{*}{Hi-C contact map prediction}  & \checkmark & $\circ$                   & \checkmark &  \textbf{0.8724} &  \textbf{0.8001} \\ \midrule
                         & \checkmark & -                         & -                         & 0.8236                                                         & 0.8054                                                         \\
                         & \checkmark & $\circ$                   & \checkmark & 0.8494                                                         & 0.8226                                                         \\
                         & -                         & \checkmark & -                         & 0.9065                                                         & 0.9072                                                         \\
                         & $\circ$                         & \checkmark & \checkmark                         & 0.9135                                                         & 0.9159                                                         \\
                         & \checkmark & \checkmark & -                         & 0.9091          & 0.8859                                                         \\
\multirow{-5}{*}{Chromatin loops dectection}  & \checkmark & \checkmark & \checkmark & \textbf{0.9209}                                                & \textbf{0.9194}                                                \\ \midrule
                         & \checkmark & -                         & -                         & 0.8514                                                         & 0.8710                                                         \\
                         & \checkmark & $\circ$                   & \checkmark & 0.8684                                                         & 0.8870                                                         \\
                         & \checkmark & \checkmark & -                         & 0.8614                                                         & 0.8755          \\
\multirow{-4}{*}{CAGE-seq expression prediction}  & \checkmark & \checkmark & \checkmark & \textbf{0.8833} & \textbf{0.9077} \\ \bottomrule
\end{tabular}}
\vspace{-3mm}
\end{wraptable}

In addition, we assess the representation learning ability of MIX-HIC across three aspects: the contribution of pre-training, the superiority of multimodal over unimodal representations, and the efficiency of modality completion. For the Hi-C contact map prediction task, the model is trained to predict the contact map using only 1D epigenomic tracks as input. The Hi-C contact map itself is the prediction target (output), making its use as an input feature methodologically invalid. Detailed comparison results appear in Table~\ref{tab:rep}. The pre-trained bimodal MIX-HIC achieves superior performance compared to its non-pre-trained counterpart, confirming pre-training's benefit for unified representations. Moreover, while pre-trained bimodal MIX-HIC consistently outperforms single-modal variants, the non-pre-trained bimodal version underperforms compared to using Hi-C data alone for K562 chromatin loop detection due to heterogeneity between bimodal data, highlighting the importance of learning both modality-specific information and general patterns. Finally, when Hi-C data is unavailable, combining epigenomic embeddings with inferred Hi-C embeddings improves performance over using epigenomic data alone, validating the cross-modal mapping block's effectiveness for data scarcity scenarios.

\vspace{-3mm}
\section{Conclusion}
\vspace{-3mm}
\label{sec:conclusion}
In this work, we present MIX-HIC, a novel multimodal foundation model for diverse 3D genome downstream tasks by integrating Hi-C contact maps and epigenomic tracks. To facilitate a unified and comprehensive understanding of 3D genome organization, we construct the largest paired 3D genome dataset to date, comprising over one million high-quality samples. As the first multimodal foundation model in this domain, MIX-HIC incorporates two key innovations: (1) a cross-modal interaction block that jointly learns modal-invariant and modal-specific representations, effectively capturing both shared and unique biological patterns across modalities; and (2) a cross-modal mapping block that regularizes the multimodal feature space and enables robust imputation of missing modalities features, alleviating the practical challenges posed by the high costs of Hi-C contact map acquisition. Comprehensive experimental results demonstrate that MIX-HIC achieves state-of-the-art performance on three key downstream tasks across two cell lines. 




\section{Acknowledgments}
This work was supported in part by the National Key R\&D Program of China (Grant No.2023YFF0725001), in part by the National Natural Science Foundation of China (Grant Nos.92370204, 62572413), in part by the guangdong Basic and Applied Basic Research Foundation (Grant Nos.2023B1515120057, 2025A1515012944), and in part by the Education Bureau of Guangzhou.

\bibliographystyle{unsrt}
\bibliography{reference}

\newpage
\appendix

\section{Implementation Details.}
\label{sec:imp}

In this section, we present the implementation details of MIX-HIC, including data source and processing, whole-chromosome loop annotation, as well as experimental settings. 


\begin{table*}[h]
\centering
\caption{Data sources and accession numbers from 4DN Data Portal and ENCODE Data Portal.}
\label{tab:ref_num}
\vskip 0.1in
\renewcommand{\arraystretch}{1.2}
\resizebox{\linewidth}{!}{
\begin{tabular}{@{}l|c|c|c|c|c@{}}
\toprule
Cell line & Hi-C         & DNase-seq   & ATAC-seq    & ChIA-PET    & CAGE-seq    \\ \midrule
GM12878   & 4DNFI1UEG1HD & ENCSR000EMT & ENCSR095QNB & ENCSR184YZV & ENCSR000CKA \\
K562      & 4DNFITUOMFUQ & ENCSR000EOT & ENCSR956DNB & ENCSR597AKG & ENCSR000CJN \\
HepG2     & 4DNFICSTCJQZ & ENCSR000EJV & ENCSR042AWH & ENCSR411IVB & -           \\
HCT116    & 4DNFIXTAS6EE & ENCSR000ENM & ENCSR872WGW & ENCSR278IZK & -           \\
IMR90     & 4DNFIH7TH4MF & ENCSR477RTP & ENCSR200OML & ENCSR076TTY & -           \\
WTC11     & 4DNFIVSCH2CH & ENCSR785ZUI & ENCSR541KFY & ENCSR353ASS & -           \\ \bottomrule
\end{tabular}}
\vspace{-3mm}
\end{table*}

\subsection{Data Source and Processing}
\label{sec:data_source}
Detailed reference numbers for all data sources are provided in Table~\ref{tab:ref_num}. We download the BAM files for epigenomic tracks and convert them into bigWig files using deepTools \cite{ramirez2016deeptools2} with RPGC normalization. Two 250 kb epigenomic regions, corresponding to the x-axis and y-axis of a Hi-C contact map, are first averaged over every 100 bp and transformed using $log(x+1)$ to reduce data variability, and then the processed regions are concatenated to form the final epigenomic tracks input with a length of 5,000. The Hi-C contact maps are binned at 5 kb resolution and normalized using KR normalization, and then divided into $50 \times 50$ sub-matrices based on the loci of each sample.

As the first attempt to create the multimodal 3D genome foundation model, we prioritize epigenomic tracks with ATAC-seq and DNase-seq because they capture the genome's foundational regulatory information. While DNase-seq and ATAC-seq have some inherent redundancy, they also offer complementary insights due to their distinct enzymatic biases. The enzymes' different cutting biases cause each assay to detect unique accessible sites, which together yield a more complete accessibility landscape \cite{meyer2014identifying}.


The training samples of three downstream tasks are generated using two widely adopted cell lines, GM12878 and K562, as described below. First, the Hi-C contact maps and epigenomic tracks within a 250 kb genomic region upstream and downstream of the gene transcription start site (TSS) are used to predict CAGE-seq expression, with TSS annotations obtained from previous work \cite{karbalayghareh2022chromatin}. For this task, the number of training samples for GM12878 and K562 are 16,046 and 14,739, respectively. Second, the same pairwise samples of Hi-C contact maps and epigenomic tracks are used for the Hi-C contact map prediction task, with the same sample sizes as mentioned above. Third, the positive chromatin loops (1,344,270 for GM12878 and 137,558 for K562) are derived from the ChIA-PET data, while an equal number of negative loops are randomly sampled based on two criteria following prior research \cite{wang2022dloopcaller}: (1) matching the distance distribution of positive samples using the distances probability density function, and (2) selecting interactions with distances greater than the maximum distance observed in the positive samples to enhance the diversity of the negative samples. Following previous work \cite{wang2022dloopcaller, tan2023cell, yang2023epiphany}, we partition the chromosomes into distinct training, validation, and test sets. Specifically, chromosomes 10 and 11 serve as the validation set, chromosomes 3, 13, and 17 as the test set, and the remaining chromosomes are used for model training across three downstream tasks.

\subsection{Whole-chromosome Loops Annotation}
\label{sec:whole_chrom}
The well-trained MIX-HIC model is capable of predicting all potential chromatin interactions across individual chromosomes. The detection of chromatin loops within each chromosome consists of two key steps: (1) scoring potential chromatin interactions using MIX-HIC and (2) aggregating loops through a clustering algorithm. First, candidate elements are identified by examining the diagonals of the raw Hi-C contact matrix, where observed interaction frequencies are statistically compared to expected values based on a Poisson distribution, retaining only elements with p-values below 0.01. Subsequently, the trained MIX-HIC model is employed to predict loop probability scores, with scores exceeding 0.5 referred to as candidate loops. These candidates are further grouped to identify significant loops using a density-based clustering algorithm following \cite{zhang2022reference}. For each candidate, a local density score and a minimum distance metric are computed. Candidates exhibiting low distance values are subsequently eliminated to minimize redundancy. Cluster centers are determined via a target-decoy search, ensuring that the false discovery rate remains below 5\%. Within each cluster, the candidate demonstrating the highest local density is designated as the representative loop.

\subsection{Experimental Settings}
\label{sec:setup}
MIX-HIC is developed using Python and PyTorch, and executed on the Ubuntu platform with a Tesla A100 GPU. MIX-HIC processes input data consisting of $50\times50$ Hi-C contact maps and epigenomic tracks with a sequence length of 5,000 bps. The architecture is composed of feature encoders, task-specific decoders, and a modality fusion block. The feature encoders and task-specific decoders are structured with four layers, each comprising two transformer encoder blocks. Notably, the epigenomic feature encoder includes an additional preprocessing stage, integrating four convolutional layers followed by max-pooling operations before the transformer blocks, to effectively process and condense long sequences. Similarly, the modality fusion block is constructed with two transformer encoder blocks, ensuring efficient integration of features across modalities. 

During the pre-training stage, MIX-HIC is configured with 500 epochs, a learning rate of 1e-5, and a batch size of 256. For the CAGE-seq expression prediction task, the predefined feature dimension $C$ and learning rate are set to 256 and 1e-4, respectively. For the prediction of Hi-C contact maps and chromatin loops, these parameters are specified as 128 and 1e-5, respectively. The number of transformer blocks $T$ in each encoder, contact map-grounded fusion block, and decoder is set to 2. Fine-tuning is conducted with a batch size of 64, utilizing the AdamW optimizer \cite{loshchilov2017decoupled} with the momentum parameters $\beta_1$ and $\beta_2$ initialized to 0.9 and 0.999, respectively. The fine-tuning process is configured with a maximum of 200 epochs, and an early stopping strategy is employed with a patience parameter of 20 to prevent overfitting.

\section{Analysis of Information Gap Between Bimodal Representations.}
\label{sec:Analysis}
Simple alignment of multimodal features can result in information loss, which may compromise the performance of downstream tasks. Under conditions of perfect feature alignment, the prediction error using aligned features is at least $\Gamma_q$ greater than that using the raw inputs. Based on \cite{dong2023simmmdg}, we generalize the theorem to the case of two modalities as Theorem~\ref{equation：theorem}.
This theorem implies that if one modality is more informative than the others (\textit{i.e.}, there exists a significant information gap), perfect alignment may lead to a substantial increase in prediction error. The underlying reason is that perfect alignment constrains the aligned features to preserve only the predictive information shared across all modalities, which may result in the loss of modal-specific information that is potentially critical for achieving accurate predictions. The proof of this theorem is presented as follows:

Consider the joint mutual information $U(\mathbf{F}^1, \mathbf{F}^2; t)$. Applying the chain rule of mutual information, we obtain:
\begin{align}
U(\mathbf{F}^1, \mathbf{F}^2; t) &= U(\mathbf{F}^1; t) + U(\mathbf{F}^2; t|\mathbf{F}^1) \\
&= U(\mathbf{F}^2; t) + U(\mathbf{F}^1; t|\mathbf{F}^2).
\end{align}
Under the condition of perfect alignment between the two features, $U(\mathbf{F}^2; t|\mathbf{F}^1) = U(\mathbf{F}^1; t|\mathbf{F}^2) =0$, which implies: 
\begin{equation}
U(\mathbf{F}^1, \mathbf{F}^2; t) = U(\mathbf{F}^1; t) = U(\mathbf{F}^2; t).
\end{equation}

By applying the data processing inequality \cite{beaudry2011intuitive}: 
\begin{equation}
U(\mathbf{F}^1; t) \leq U(z_1; t); U(\mathbf{F}^2; t) \leq U(z_2; t),
\end{equation}
we derive $U(\mathbf{F}^1, \mathbf{F}^2; t)$ as follows:
\begin{align*}
U(\mathbf{F}^1, \mathbf{F}^2; t) &= \min\{U({F}^1; t), U({F}^2; t)\} \\
&\leq \min\{U(z_1; t), U(z_2; t)\} \\
&\leq \max\{U(z_1; t), U(z_2; t)\} \\
&\leq U(z_1, z_2; t).
\end{align*}

According to the variational form of conditional entropy $H(t | z_1, z_2) = \inf_h \mathbb{E}_q[\mathcal{L}_{\text{CE}}(h(z_1, z_2), t)]$ \cite{zhao2022fundamental} and the definition of mutual information $U(X;Y) = H(Y) - H(X|Y)$, we can conclude the theorem as follows:
\begin{align*}
\inf_h \mathbb{E}_q[\mathcal{L}_{\text{CE}}(h(\mathbf{F}^1, \mathbf{F}^2), t)] &- \inf_{h'} \mathbb{E}_q[\mathcal{L}_{\text{CE}}(h'(z_1, z_2), t)] \\
&= H(t | \mathbf{F}^1, \mathbf{F}^2) - H(t | z_1, z_2)\\
&= H(t) - U(\mathbf{F}^1, \mathbf{F}^2|t) - (H(t) - U(z_1, z_2|t))\\
&= U(z_1, z_2; t) - U(\mathbf{F}^1, \mathbf{F}^2; t) \\
&\geq \Gamma_q.
\end{align*}

\vspace{-2mm}
Therefore, although perfect alignment of bimodal features achieves consistency, it risks losing critical modal-specific information, potentially leading to higher prediction errors, especially when the information gap between modalities is substantial, as seen in the case of Hi-C contact maps and epigenomic tracks. To solve this problem, we introduce cross-modal interaction and cross-modal mapping blocks to capture both modal-invariant and modal-specific features, enabling a more comprehensive representation of bimodal data. The effectiveness of this approach is validated through extensive ablation studies.

\section{Additional Experimental Results.}

\subsection{Proportion of Validated Loops versus Predicted Loops}
\label{sec:proportion}
We define the quantitative metric `proportion' as the ratio of experimentally validated loops to computationally predicted loops \cite{zhang2022reference}. Table~\ref{tab:num-loops} presents a comparative analysis of chromatin interaction loops identified through ChIA-PET experiments and those predicted by different approaches. MIX-HIC exceeds runner-up methods by 28\% and 9\% proportion on GM12878 and K562 datasets, respectively.

\begin{table*}[h]
\vspace{-3mm}
\centering
\caption{Comparison of ChIA-PET-supported loops with those identified by various deep learning methods.}
\label{tab:num-loops}
\hypertarget{tab:num-loops}{}
\vskip 0.1in
\renewcommand{\arraystretch}{1.2}
\resizebox{\linewidth}{!}{
\begin{tabular}{l|ccc|ccc}
\toprule
\multirow{2}{*}{Methods} & \multicolumn{3}{c|}{GM12878}                   & \multicolumn{3}{c}{K562}                       \\ \cmidrule{2-7} 
                         & Validated Loops & Predicted Loops & Proportion & Validated Loops & Predicted Loops & Proportion \\ \midrule
HiCExplorer              & 1358            & 2688            & 0.51       & 765             & 1450            & 0.53       \\
ChromoSight              & 3845            & 6044            & 0.64       & 993            & 1628            & 0.61       \\
Peakachu                 & 1992            & 2904            & 0.69       & 795             & 1492            & 0.53       \\
DLoopCaller              & 4301            & 6878            & 0.62       & 1048            & 2274            & 0.46       \\
\rowcolor{blue!10}
MIX-HIC-Bimodal (Ours)                  & 5179            & 5893            & 0.88       & 1064            & 1588            & 0.67       \\ \bottomrule
\end{tabular}}
\vspace{-4mm}
\end{table*}

\subsection{Example of Whole-Chromosome Chromatin Loop Comparison}
\label{sec:example_loop}
\begin{figure*}[t]
\begin{center}
\centerline{\includegraphics[width=0.9\linewidth]{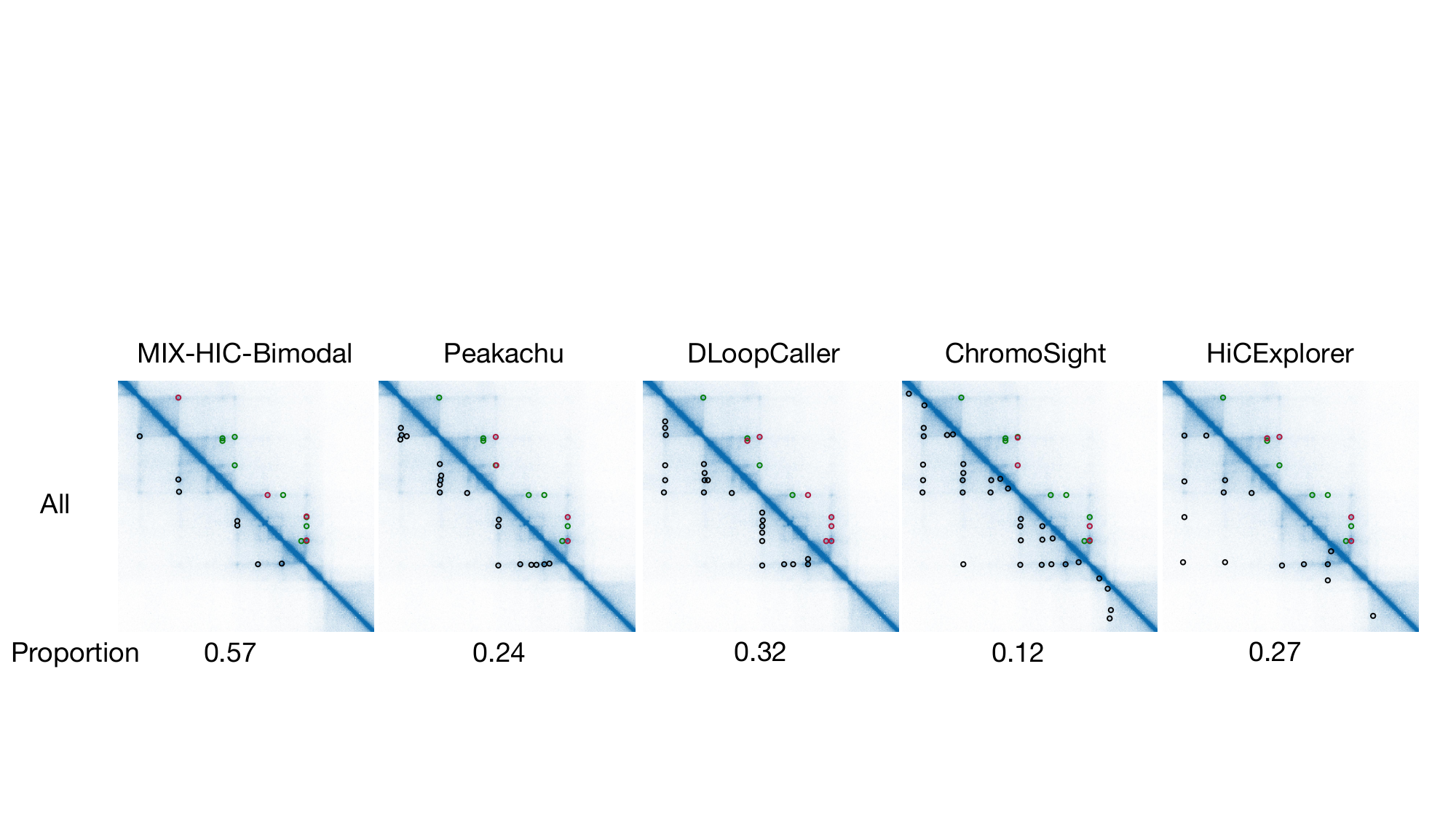}}
\caption{Comparison of predicted chromatin loops (black points, lower left) and corresponding experimentally validated ChIA-PET interactions (red points, upper right) in the chromosome 13 region (20Mb-22Mb) from the GM12878 dataset. Unmatched experimentally validated ChIA-PET interactions are represented by green points in the upper right.}
\label{fig:loop_compared}
\hypertarget{fig:loop_compared}{}
\end{center}
\end{figure*}

Figure~\ref{fig:loop_compared} illustrates the performance evaluation of chromatin loops predicted by various methods (black points, lower left) against corresponding experimentally validated interactions (red points, upper right).  Specifically, we use the same genome regions, chromosome 13: 20Mb-22Mb from the GM12878 dataset, and identical ground truth data for evaluating all compared methods. Our predictions (the first left panel) demonstrate significantly fewer false positive loops compared to experimentally validated ChIA-PET interactions, as evidenced by the correspondent counts between red and black points. Our method achieves a proportion of 0.57, compared to the runner-up's 0.32, highlighting MIX-HIC's superior ability to accurately identify chromatin loops while producing fewer false positives.

\subsection{Hyperparameter Analysis}
\label{sec:hyperparameter}
\begin{figure*}[h]
\begin{center}
\centerline{\includegraphics[width=0.9\linewidth]{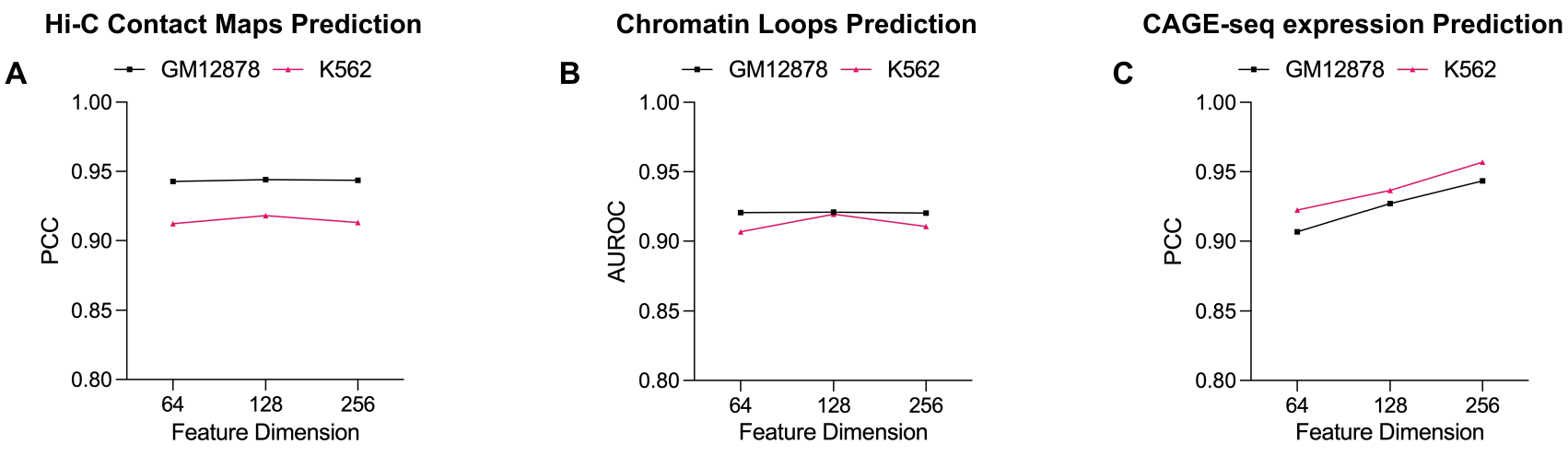}}
\caption{The performance across different predefined feature dimensions $C=\{64, 128, 256\}$ for three downstream tasks in GM12878 and K562 cell lines.}
\label{fig:dim}
\hypertarget{fig:dim}{}
\end{center}
\end{figure*}

\begin{figure*}[h]
\begin{center}
\centerline{\includegraphics[width=0.9\linewidth]{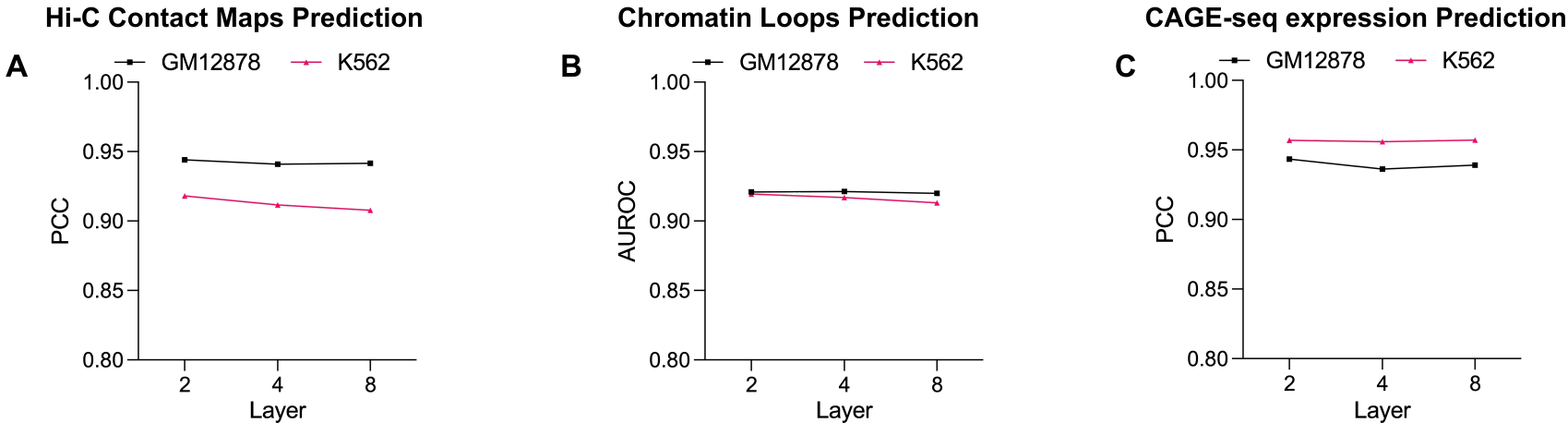}}
\caption{The performance across different Transformer layers $T=\{2, 4, 8\}$ for three downstream tasks in GM12878 and K562 cell lines.}
\label{fig:layer}
\hypertarget{fig:layer}{}
\end{center}
\end{figure*}

The number of Transformer $T$ blocks in each layer of the encoder, contact map-grounded fusion block, and decoder, along with the predefined feature dimension $C$ play a pivotal role in MIX-HIC. We evaluate the impact of this parameter by testing $C = \{64, 128, 256\}$ and $T=\{2, 4, 8\}$ as depicted in Figure~\ref{fig:dim} and Figure~\ref{fig:layer}, respectively. The highest performance is achieved at  $C=128$ for HI-C contact maps and chromatin loops prediction tasks, while $C=256$ performs best for CAGE-seq expression prediction. Additionally, $T=2$ demonstrates the most robust performance across all three downstream tasks. MIX-HIC exhibits strong resilience to parameter variations, demonstrating our model's robustness and parameter efficiency. The performance suggests our architecture reaches a ``sweet spot'' with a moderate parameter count and computational efficiency, sufficient to capture the essential biological patterns without incurring the high risk of overfitting during fine-tuning on smaller, task-specific datasets. As our pre-training corpus expands, we anticipate larger architectures will become beneficial.
These experimental results substantiate the parameter configurations employed in MIX-HIC.

\subsection{Orthogonal Constraint Enhances Feature Diversity}
\label{sec:inner}
Theorem~1 demonstrates that rigorous alignment can be detrimental to performance. The orthogonal constraint effectively mitigates this issue by promoting diversity between modality-invariant and modality-specific features, rather than enforcing strict non-overlap. To validate feature diversity, we compute the inner product between modality-invariant $\hat{X_{M}^{I}}$ and modality-specific $\hat{X_{M}^{S}}$ features of Hi-C contact maps, as well as between their counterparts $\hat{X_{E}^{I}}$ and $\hat{X_{E}^{S}}$ in epigenomic tracks on the datasets of the CAGE-seq expression prediction task. As summarized in Table~\ref{tab:inner_product}, the inner products under orthogonal constraints are orders of magnitude smaller than those without constraints (\textit{e.g.}, $1 \mathrm{e}{-5}$ versus $1.077$ for Hi-C contact maps on GM12878 dataset). Our analysis of inner product results between these embeddings confirms that the constraint successfully generates near-orthogonal representations.

\begin{table}[h]
\centering
\caption{Inner products between modality-invariant and modality-specific features.}
\label{tab:inner_product}
\vskip 0.1in
\renewcommand{\arraystretch}{1.2}
\begin{tabular}{l|cc|cc}
\toprule
\multirow{2}{*}{Dataset} & \multicolumn{2}{c|}{With Pre-training}   & \multicolumn{2}{c}{Without Pre-training} \\ \cline{2-5} 
                         & Hi-C contact maps & Epigenomic tracks & Hi-C contact maps  & Epigenomic tracks  \\ \midrule
GM12878                  & $0.003 \pm 3 \mathrm{e}{-4}$    & $0.002 \pm 0.001$   & $1.419 \pm 0.095$     & $0.164 \pm 0.059$    \\
K562                     & $1 \mathrm{e}{-5} \pm 4 \mathrm{e}{-4}$     & $0.002 \pm 0.002$   & $1.077 \pm 0.074$    & $0.144 \pm 0.057$    \\ \bottomrule
\end{tabular}
\end{table}

\subsection{In Silico Perturbation Validates Biological Grounding of Chromatin Loop Detection}
\label{sec:interpre}
The presence and strength of certain epigenomic signals are highly associated with the formation of chromatin loops \cite{zhao2019chromatin, sahin2021hic}. For example, CTCF binding sites are typically located within regions of open chromatin, identifiable as peaks in assays such as DNase-seq or ATAC-seq.

To validate that MIX-HIC learns the fundamental principle linking epigenomic signals to 3D genome architecture, we design an in \textit{silico} perturbation experiment. We hypothesize that the model's loop detection should be governed by the underlying epigenomic signals at the loop anchors. We focus our analysis on 118 high-confidence chromatin loops in the K562 cell line, each characterized by convergent CTCF motifs at their anchors (identified via FIMO \cite{grant2011fimo}). We then systematically attenuate the input epigenomic tracks (ATAC-seq and DNase-seq) by down-sampling the signal intensity of peaks within these anchor regions at varying ratios. These perturbed epigenomic profiles are then fed into the MIX-HIC-InferMap model (trained on GM12878) to assess the impact on loop detection. The results are shown in Table~\ref{tab:interpre}.

Using the unaltered epigenomic data, MIX-HIC successfully recalls all 118 CTCF-mediated loops. As we progressively degrade the epigenomic signals at the loop anchors, the model’s recall for these loops decreases. This demonstrates MIX-HIC's predictions are mechanistically grounded in biologically pertinent epigenomic features. This result demonstrates the model's biological interpretability, confirming the capture of a fundamental principle of genome organization.

\begin{table}[h]
\centering
\caption{Impact of Epigenomic Signal Attenuation on MIX-HIC Loop Recall.}
\label{tab:interpre}
\vskip 0.1in
\renewcommand{\arraystretch}{1.2}
\begin{tabular}{@{}lccccc@{}}
\toprule
\textbf{Varying ratio} & 0.0 & 0.5 & 0.7 & 0.8 & 0.9 \\ \midrule
\textbf{MIX-HIC-InferMap} & 100\% (118) & 98\% (116) & 61\% (72) & 15\% (18) & 0\% (0) \\ \bottomrule
\end{tabular}
\end{table}

\subsection{Robustness Analysis Under Noisy Conditions}
The inherent noise and sparsity of Hi-C data \cite{yardimci2019measuring, spill2019binless} present a critical challenge for developing robust and generalizable genomic models. To systematically assess the robustness of MIX-HIC, we conduct a controlled experiment simulating low-coverage and noisy scenarios. We corrupt Hi-C contact maps by perturbing different ratios of non-zero contacts with sparsity and Gaussian noise. Under these varying noise levels, MIX-HIC and the supervised baseline Peakachu are evaluated on the chromatin loop detection task using the K562 cell line dataset.

As shown in Table~\ref{tab:robustness_analysis}, the performance of Peakachu degrades sharply as noise increases, falling to near-random performance (0.5091 AUROC) when 70\% of the contacts are disturbed. In contrast, MIX-HIC exhibits remarkable robustness, with its performance declining by less than 7\% around all noise ratios.

We attribute this resilience to the power of our pre-training paradigm, which learns the fundamental principles of 3D genome organization from over 1 million samples. This experiment demonstrates that MIX-HIC develops a robust biological representation and is thus particularly suitable for analyzing noisy or low-coverage datasets. We will add this analysis to our manuscript to further strengthen the paper.

\begin{table}[h]
\centering
\caption{AUROC comparison of MIX-HIC and Peakachu on chromatin loop detection under varying levels of simulated noise and sparsity.}
\label{tab:robustness_analysis}
\vskip 0.1in
\renewcommand{\arraystretch}{1.2}
\begin{tabular}{lcccc}
\toprule
\textbf{Varying ratio} & 0.0 & 0.5 & 0.7 & 0.9 \\
\midrule
Peakachu & 0.8833 & 0.7659 & 0.5091 & - \\
MIX-HIC  & 0.9194 & 0.8899 & 0.8754 & 0.8486 \\
\bottomrule
\end{tabular}
\end{table}

\section{Baseline Methods and Assets}
\label{sec:assets}
In this study, the effectiveness of MIX-HIC is evaluated on three downstream tasks, including Hi-C contact map prediction, chromatin loop detection, and CAGE-seq expression prediction. We employ four kinds of methods involving Epiphany \cite{yang2023epiphany}, C.Origami \cite{tan2023cell}, EPCOT-LSTM \cite{zhang2023generalizable}, and EPCOT-Transformer \cite{zhang2023generalizable} for comparison on the Hi-C contact map prediction task. For chromatin loop detection task, two statistical-based methods (ChromoSight \cite{matthey2020computer} and HiCExplorer \cite{wolff2020galaxy}), as well as two supervised learning-based approaches (Peakachu \cite{salameh2020supervised} and DLoopCaller \cite{wang2022dloopcaller}) are utilized for evaluation. The CAGE-seq expression prediction involves EPI-CNN \cite{karbalayghareh2022chromatin}, EPI-Graph \cite{karbalayghareh2022chromatin}, EPCOT-LSTM \cite{zhang2023generalizable}, and EPCOT-Transformer \cite{zhang2023generalizable} to compare with MIX-HIC. More details of these baselines are described below.
\begin{itemize}[leftmargin=10pt]
\item Epiphany \cite{yang2023epiphany} predicts cell-type-specific Hi-C contact maps using bidirectional LSTMs to encode epigenomic tracks.
\item C.Origami \cite{tan2023cell} introduces a multimodal framework that predicts chromatin organization from DNA sequence, CTCF binding signals, and epigenomic tracks, enabling the effective identification of regulatory elements.
\item ChromoSight \cite{matthey2020computer} proposes a computer vision-inspired algorithm for chromatin loop detection that employs expert-defined pattern templates, demonstrating computational efficiency across diverse species without requiring training data.
\item HiCExplorer \cite{wolff2020galaxy} identifies significant chromatin interactions by analyzing Hi-C contact matrices, employing binomial distribution modeling to distinguish true loops from background noise while controlling for distance-dependent contact probability.
\item Peakachu \cite{salameh2020supervised} develops a supervised random forest classification framework that leverages chromatin interaction labels to predict chromatin loops from Hi-C contact maps, outperforming statistical enrichment methods in identifying short-range interactions.
\item Xu~et~al.~\cite{wang2022dloopcaller} propose DLoopCaller, a supervised deep learning method that predicts genome-wide chromatin loops by integrating epigenomic tracks with raw Hi-C contact maps.
\item Karbalayghareh~et~al.~\cite{karbalayghareh2022chromatin} extracts local features from epigenomic tacks using convolutional neural networks (EPI-CNN) and incorporates Hi-C contact maps through graph attention networks (EPI-Graph) to predict CAGE-seq expression.
\item EPCOT \cite{zhang2023generalizable} introduces a pre-training and fine-tuning deep learning method that leverages Transformer (EPCOT-Transformer) or LSTM (EPCOT-LSTM) architectures to predict Hi-C contact maps and CAGE-seq expression profiles from epigenomic tracks and DNA sequences, achieving generalizable representations across diverse cell types.
\end{itemize}

The methods Epiphany, C.Origami, ChromoSight, Peakachu, and DLoopCaller are licensed under a Creative Commons Attribution 4.0 International License (CC BY 4.0), while EPI-CNN, EPI-Graph, EPCOT-Transformer, and EPCOT-LSTM are available under an Attribution-NonCommercial 4.0 International License (CC BY-NC 4.0). HiCExplorer is distributed under the GNU General Public License v3.0 (GPL-3.0).
The licenses of other open-source tools utilized in this work are summarized in Table~\ref{tab:license}.

\begin{table}[h]
\centering
\caption{License of softwares used in this study.}
\label{tab:license}
\vskip 0.1in
\renewcommand{\arraystretch}{1.2}
\begin{tabular}{ccc}
\toprule
Software     & License            & URL \\ \midrule
Juicer       & MIT license        & \url{https://github.com/aidenlab/juicer} \\
pyBigWig     & MIT license        & \url{https://github.com/deeptools/pyBigWig} \\
hicstraw     & MIT license        & \url{https://github.com/aidenlab/straw} \\
Huggingface  & Apache-2.0         & \url{https://huggingface.co/} \\
Scikit-Learn & BSD-3-Clause       & \url{https://scikit-learn.org/stable/} \\
Numpy        & BSD-3-Clause       & \url{https://numpy.org/} \\
Pytorch      & BSD-3-Clause       & \url{https://pytorch.org/} \\
Matplotlib   & Matplotlib License & \url{https://matplotlib.org/} \\ \bottomrule
\end{tabular}
\vspace{-3mm}
\end{table}

\section{Broader Impacts and Limitations} 
\label{sec:broader_impact}
\textbf{Broader Impacts.}
The three-dimensional chromatin architecture fundamentally governs both cellular differentiation and disease progression by mediating genomic interactions. This necessitates the development of Hi-C foundation models to systematically unravel the mechanistic basis of gene regulatory networks in both physiological and disease contexts. MIX-HIC fundamentally advances 3D genomics analysis by addressing two critical limitations of current approaches. First, they are typically designed for single tasks with limited cross-task knowledge transfer capability. Second, they predominantly rely on single-modality data (either Hi-C or epigenomic tracks alone) due to the scarcity of paired multimodal datasets, resulting in an incomplete understanding of chromatin organization. 

To address these challenges, we develop MIX-HIC, the first multimodal foundation model for 3D genomics, with three key innovations: 1) We curate the largest paired dataset (over 1 million samples) of Hi-C and epigenomic tracks to overcome data scarcity. 2) Our novel cross-modal interaction and mapping blocks simultaneously capture both modality-invariant and modality-specific features, enabling a reliable complement of missing modality features. 3) As a foundation model, MIX-HIC enables versatile adaptation to diverse downstream tasks, facilitating knowledge transfer across different 3D genomics tasks where current approaches operate in isolation. In summary, MIX-HIC provides a universal computational platform for systematically deciphering the coordinated mechanisms between chromatin spatial organization and epigenetic regulation, while pioneering new avenues for multimodal data integration in disease mechanism research, precision medicine, and synthetic biology applications.

\textbf{Limitations.} 
Although MIX-HIC has exhibited promising performance, several areas warrant further improvement. Current methods in the 3D genome field for processing long-range DNA sequences are often time-intensive. In future work, MIX-HIC could further enhance its capabilities by integrating DNA sequence information through leveraging recent advancements like MambaDNA \cite{schiff2024caduceus} and HyenaDNA \cite{nguyen2024hyenadna} to facilitate feature extraction of genomic sequences. 
Additionally, while the current version of MIX-HIC serves as a bulk-seq level foundation model, developing a single-cell version through the integration of large-scale multimodal single-cell data would be valuable to effectively address the inherent sparsity and noise in single-cell analyses. 
Finally, the diversity of cell lines in our pre-training set is an area for future enhancement. Expanding the pre-training corpus by integrating the paired Hi-C and epigenomic datasets available from ENCODE will broaden MIX-HIC's applicability across a wider range of biological contexts. Overall, this study presents MIX-HIC, a versatile foundation model that integrates 3D genome structures with chromatin accessibility, providing an efficient framework for advancing genomic organization research and related fields.

\clearpage
\newpage
\section*{NeurIPS Paper Checklist}

\begin{enumerate}

\item {\bf Claims}
    \item[] Question: Do the main claims made in the abstract and introduction accurately reflect the paper's contributions and scope?
    \item[] Answer: \answerYes{} 
    \item[] Justification: The contributions of this work are summarized in the introduction section, while the score is clearly described in the abstract.

\item {\bf Limitations}
    \item[] Question: Does the paper discuss the limitations of the work performed by the authors?
    \item[] Answer: \answerYes{} 
    \item[] Justification: The limitations are included in Appendix \ref{sec:broader_impact}.

\item {\bf Theory assumptions and proofs}
    \item[] Question: For each theoretical result, does the paper provide the full set of assumptions and a complete (and correct) proof?
    \item[] Answer: \answerYes{} 
    \item[] Justification: The proofs of the theorems are detailed in Appendix \ref{sec:Analysis}.

    \item {\bf Experimental result reproducibility}
    \item[] Question: Does the paper fully disclose all the information needed to reproduce the main experimental results of the paper to the extent that it affects the main claims and/or conclusions of the paper (regardless of whether the code and data are provided or not)?
    \item[] Answer: \answerYes{} 
    \item[] Justification: The implementation details and experimental settings are provided in Appendix \ref{sec:imp}.

\item {\bf Open access to data and code}
    \item[] Question: Does the paper provide open access to the data and code, with sufficient instructions to faithfully reproduce the main experimental results, as described in supplemental material?
    \item[] Answer: \answerYes{} 
    \item[] Justification: All data used in this work are publicly available, as detailed in Appendix~\ref{sec:imp}. The code of MIX-HIC is available in our GitHub.

\item {\bf Experimental setting/details}
    \item[] Question: Does the paper specify all the training and test details (e.g., data splits, hyperparameters, how they were chosen, type of optimizer, etc.) necessary to understand the results?
    \item[] Answer: \answerYes{} 
    \item[] Justification: All the training hyperparameters are provided in Appendix~\ref{sec:setup}. We also perform hyperparameter analysis for two critical hyperparameters in Appendix~\ref{sec:hyperparameter}.

\item {\bf Experiment statistical significance}
    \item[] Question: Does the paper report error bars suitably and correctly defined or other appropriate information about the statistical significance of the experiments?
    \item[] Answer: \answerYes{} 
    \item[] Justification: We report error bars for the few-shot learning experiments in Figure~\ref{fig:fewshot}, computed over five independent runs with varying random seeds.

\item {\bf Experiments compute resources}
    \item[] Question: For each experiment, does the paper provide sufficient information on the computer resources (type of compute workers, memory, time of execution) needed to reproduce the experiments?
    \item[] Answer: \answerYes{} 
    \item[] Justification: We provide the computer resources used in this work in Appendix~\ref{sec:setup}.
    
\item {\bf Code of ethics}
    \item[] Question: Does the research conducted in the paper conform, in every respect, with the NeurIPS Code of Ethics \url{https://neurips.cc/public/EthicsGuidelines}?
    \item[] Answer: \answerYes{} 
    \item[] Justification: This work follows the NeurIPS Code of Ethics.

\item {\bf Broader impacts}
    \item[] Question: Does the paper discuss both potential positive societal impacts and negative societal impacts of the work performed?
    \item[] Answer: \answerYes{} 
    \item[] Justification: We state the broader impacts of this work in Appendix~\ref{sec:broader_impact}.
    
\item {\bf Safeguards}
    \item[] Question: Does the paper describe safeguards that have been put in place for responsible release of data or models that have a high risk for misuse (e.g., pretrained language models, image generators, or scraped datasets)?
    \item[] Answer: \answerNA{} 
    \item[] Justification: This work poses no such risks and the model is free to share.

\item {\bf Licenses for existing assets}
    \item[] Question: Are the creators or original owners of assets (e.g., code, data, models), used in the paper, properly credited and are the license and terms of use explicitly mentioned and properly respected?
    \item[] Answer: \answerYes{} 
    \item[] Justification: We cite all referenced sources appropriately, which are provided in Appendix~\ref{sec:assets}.

\item {\bf New assets}
    \item[] Question: Are new assets introduced in the paper well documented and is the documentation provided alongside the assets?
    \item[] Answer: \answerNA{} 
    \item[] Justification: This paper does not introduce new assets.

\item {\bf Crowdsourcing and research with human subjects}
    \item[] Question: For crowdsourcing experiments and research with human subjects, does the paper include the full text of instructions given to participants and screenshots, if applicable, as well as details about compensation (if any)? 
    \item[] Answer: \answerNA{} 
    \item[] Justification: All datasets used in this work stem from humans are anonymized, and sourced from publicly available publications to ensure privacy compliance. The source publications address key considerations regarding human subjects research.

\item {\bf Institutional review board (IRB) approvals or equivalent for research with human subjects}
    \item[] Question: Does the paper describe potential risks incurred by study participants, whether such risks were disclosed to the subjects, and whether Institutional Review Board (IRB) approvals (or an equivalent approval/review based on the requirements of your country or institution) were obtained?
    \item[] Answer: \answerNA{} 
    \item[] Justification: All datasets used in this work stem from humans are anonymized, and sourced from publicly available publications to ensure privacy compliance. The source publications address key considerations regarding human subjects research.

\item {\bf Declaration of LLM usage}
    \item[] Question: Does the paper describe the usage of LLMs if it is an important, original, or non-standard component of the core methods in this research? Note that if the LLM is used only for writing, editing, or formatting purposes and does not impact the core methodology, scientific rigorousness, or originality of the research, declaration is not required.
    \item[] Answer: \answerNA{} 
    \item[] Justification: The LLM tools are only used for writing.

\end{enumerate}

\end{document}